\documentclass[10pt,journal]{IEEEtran}

\usepackage[pdftex]{graphicx}
\DeclareGraphicsExtensions{.pdf,.jpg,.gif,.png}

\usepackage{multirow,setspace,verbatim,amsfonts,graphicx,amsmath,amsthm,amsbsy,amssymb,epsfig,url,cite}
\usepackage{amsthm}
\usepackage{gensymb}
\usepackage{graphicx}

\usepackage{amsfonts}
\usepackage{amsmath,bm}
\usepackage{amssymb}
\usepackage{amsthm}
\usepackage{tikz,graphicx}
\usepackage{mathrsfs}

\usepackage{algorithmicx}

\usepackage{array}
\usepackage{booktabs}

\usepackage{amsfonts}
\usepackage{amsmath,bm}
\usepackage{amssymb}
\usepackage{amsthm}
\usepackage{tikz,graphicx}
\usepackage{mathrsfs} 
\usepackage[algo2e]{algorithm2e} 

\usepackage{algorithm}
\usepackage{array,multirow}
\usepackage{color}
\usepackage{caption}
\usepackage{subcaption}

\usepackage{subfiles}
\usepackage{titling}

\newcommand{\bk}{\mathbf{k}}

\newcommand{\bu}{\mathbf{u}}
\newcommand{\bv}{\mathbf{v}}

\newcommand{\bz}{\mathbf{z}}

\newcommand{\NN}{\mathbb{N}}
\newcommand{\RR}{\mathbb{R}}

\newcommand{\Ab}{{\mathbf A}}
\newcommand{\Bb}{{\mathbf B}}
\newcommand{\Cb}{{\mathbf{C}}}
\newcommand{\Db}{{\mathbf D}}
\newcommand{\Eb}{{\mathbf E}}

\newcommand{\Ib}{{\mathbf I}}

\newcommand{\Mb}{{\mathbf M}}

\newcommand{\Pb}{{\mathbf P}}

\newcommand{\Sb}{{\mathbf S}}
\newcommand{\Tb}{{\mathbf T}}
\newcommand{\Ub}{{\mathbf U}}
\newcommand{\Vb}{{\mathbf V}}

\newcommand{\Xb}{{\mathbf X}}
\newcommand{\Yb}{{\mathbf Y}}
\newcommand{\Zb}{{\mathbf Z}}

\newcommand{\bb}{{\mathbf b}}

\newcommand{\fb}{{\mathbf f}}

\newcommand{\hb}{{\mathbf h}}

\newcommand{\kb}{{\mathbf k}}

\newcommand{\rb}{{\mathbf r}}

\newcommand{\vb}{{\mathbf v}}

\newcommand{\xb}{{\mathbf x}}
\newcommand{\yb}{{\mathbf y}}
\newcommand{\zb}{{\mathbf z}}

\newcommand{\Phib}{{\boldsymbol {\Phi}}}
\newcommand{\Psib}{{\boldsymbol {\Psi}}}

\newcommand{\Lambdab}{{\boldsymbol {\Lambda}}}
\newcommand{\Upsilonb}{{\boldsymbol {\Upsilon}}}
\newcommand{\Sigmab}{{\boldsymbol {\Sigma}}}
\newcommand{\Rd}{{\mathbb R}}
\newcommand{\Cd}{{\mathbb C}}

\newcommand{\psib}{{\boldsymbol{\psi}}}

\newcommand{\Ybc}{{\boldsymbol{\mathcal Y}}}

\newcommand{\Zbc}{{\boldsymbol{\mathcal Z}}}

\newcommand{\Hbc}{{\boldsymbol{\mathcal H}}}
\newcommand{\Xbc}{{\boldsymbol{\mathcal X}}}

\newcommand{\rank}{\textsc{rank}}
\newcommand{\hank}{\mathbb{H}}

\newcommand{\Rbc}{{\mathfrak{R}}}

\newcommand{\Bc}{{\mathcal B}}

\newcommand{\Dc}{{\mathcal D}}
\newcommand{\Ec}{{\mathcal E}}
\newcommand{\Fc}{{\mathcal F}}

\newcommand{\Pc}{{\mathcal P}}

\newcommand{\Tc}{{\mathcal T}}

\newcommand{\re}{{\mathrm{Re}}}
\newcommand{\im}{{\mathrm{Im}}}

\newcommand{\oneb}{\boldsymbol{1}}

\newtheorem{theorem}{Theorem}

\usepackage{xcolor}

\newcommand{\beginsupplement}{%
        \setcounter{table}{0}
        \renewcommand{\thetable}{S\arabic{table}}%
        \setcounter{figure}{0}
        \renewcommand{\thefigure}{S\arabic{figure}}%
     }

\begin{document}

\title{ $k$-Space   Deep Learning for Accelerated MRI }
\date{\vspace{-4ex}}

\author{Yoseob~Han,~
        Leonard Sunwoo,~
        and~Jong~Chul~Ye,~\IEEEmembership{Senior Member,~IEEE}
\thanks{Y. Han and J.C. Ye with the Department of Bio and Brain Engineering, Korea Advanced Institute of Science and Technology (KAIST), 
		Daejeon 34141, Republic of Korea (e-mail: \{hanyoseob,jong.ye\}@kaist.ac.kr). L. Sunwoo is with the Department of Radiology, Seoul National University College of Medicine, Seoul National University Bundang Hospital, Seongnam, Republic of Korea.
		J.C. Ye is also with the Department of Mathematical Sciences, KAIST.} 
\thanks{This work is supported by Korea Science and Engineering Foundation, Grant
		number NRF2016R1A2B3008104.}
}


\maketitle

\begin{abstract}
The annihilating filter-based low-rank Hankel matrix approach (ALOHA) is one of the state-of-the-art compressed sensing approaches that directly interpolates the missing $k$-space data using low-rank Hankel matrix completion. The success of ALOHA is due to the concise signal representation in the $k$-space domain thanks to the duality between structured low-rankness in the $k$-space domain and the image domain sparsity. Inspired by the recent mathematical discovery  that links convolutional neural networks to Hankel matrix decomposition using data-driven framelet basis,
here we  propose  a fully data-driven  deep learning algorithm for $k$-space interpolation. Our network can be also easily applied to non-Cartesian $k$-space trajectories by simply adding an additional regridding layer. Extensive numerical experiments show that the proposed deep learning method consistently outperforms  the existing image-domain deep learning approaches.
\end{abstract}

\begin{IEEEkeywords}
Compressed sensing MRI,  Deep Learning,  Hankel structured low-rank completion,  Convolution framelets
\end{IEEEkeywords}

\IEEEpeerreviewmaketitle

\section{Introduction}
\label{sec:introduction}

\IEEEPARstart{R}{ecently}, 
inspired by the tremendous success of deep learning \cite{krizhevsky2012imagenet,he2016deep,ronneberger2015u}, 
many researchers have investigated deep learning approaches  for MR reconstruction problems
and successfully demonstrated significant performance gain
\cite{wang2016accelerating,hammernik2018learning,lee2018deep,han2017deep,jin2017deep,schlemper2018deep,zhu2018image}.

 In particular,
Wang et al \cite{wang2016accelerating}  used the deep learning reconstruction either as  an initialization  or  a  regularization term.
Deep network architecture using unfolded iterative compressed sensing (CS) algorithm was also proposed   to learn a set of  regularizers and associated filters
\cite{hammernik2018learning}. 
 These works were followed by novel
extension using deep residual learning \cite{lee2018deep}, domain adaptation \cite{han2017deep}, data consistency layers \cite{schlemper2018deep}, etc. 
An extreme form of the neural network called Automated Transform by Manifold Approximation
(AUTOMAP) \cite{zhu2018image} even attempts to estimate the Fourier transform itself using fully connected layers.
All these pioneering works have consistently demonstrated superior reconstruction performances over the compressed sensing approaches \cite{lustig2007sparse,jung2009k,shin2014calibrationless,haldar2014low,jin2016general,ongie2016off}
at significantly lower run-time computational complexity.

\begin{figure}[!hbt] 	
\centerline{\includegraphics[width=0.65\linewidth]{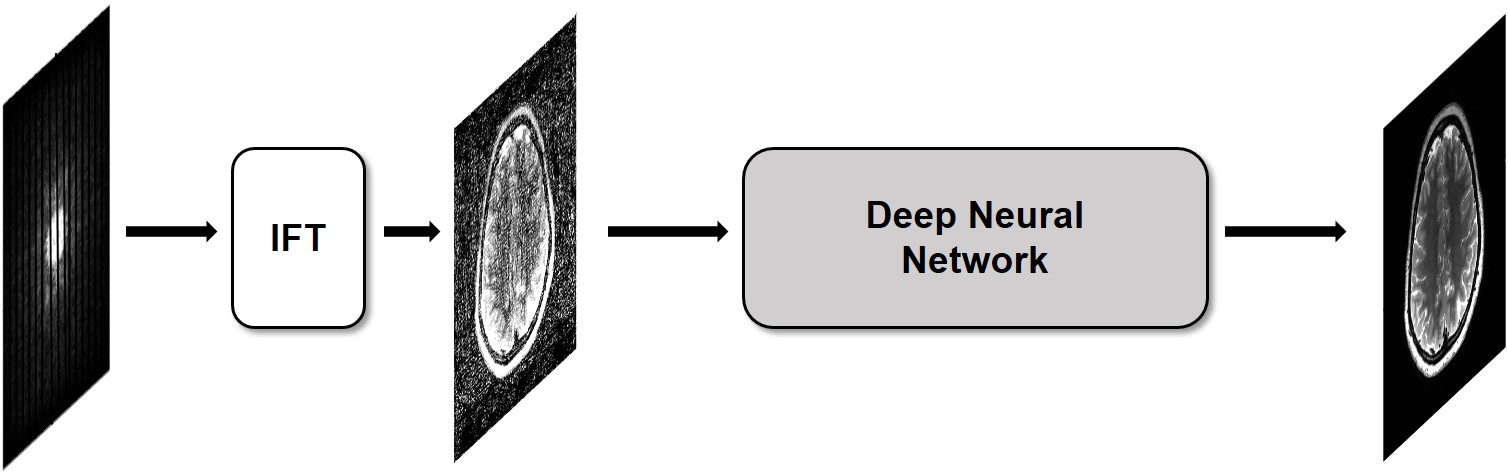}}
\vspace{-0.4cm}
\centerline{\mbox{(a)}}
\vspace{0.5cm}
\centerline{\includegraphics[width=0.65\linewidth]{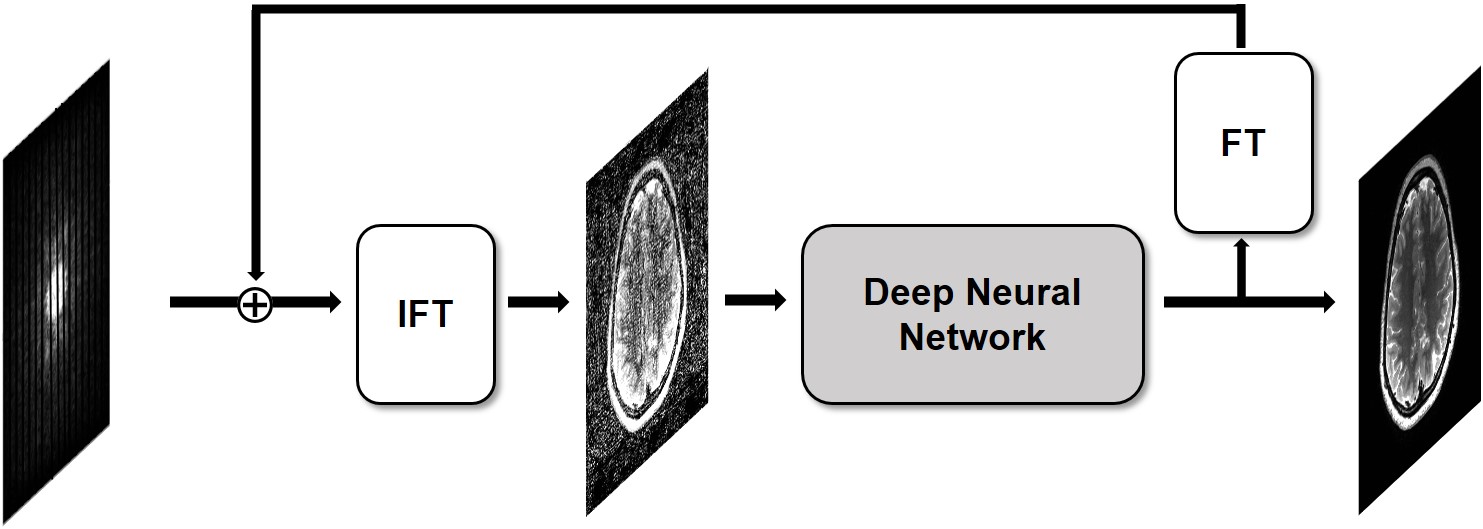}}
\vspace{-0.4cm}
\centerline{\mbox{(b)}}
\centerline{\includegraphics[width=0.65\linewidth]{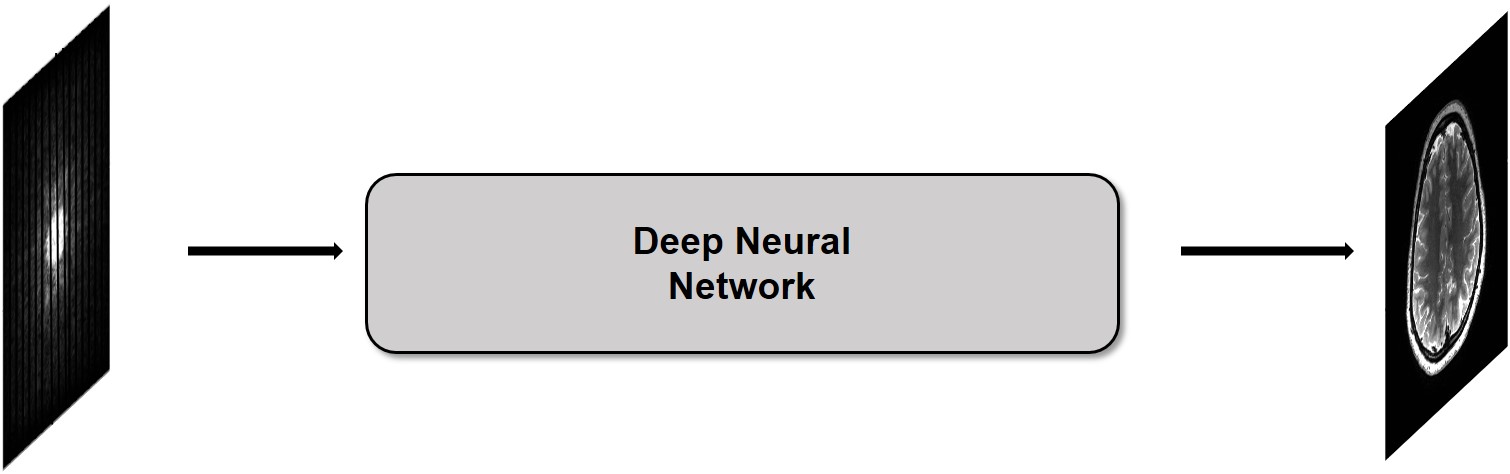}}
\vspace{-0.4cm}
\centerline{\mbox{(c)}}
\centerline{\includegraphics[width=0.65\linewidth]{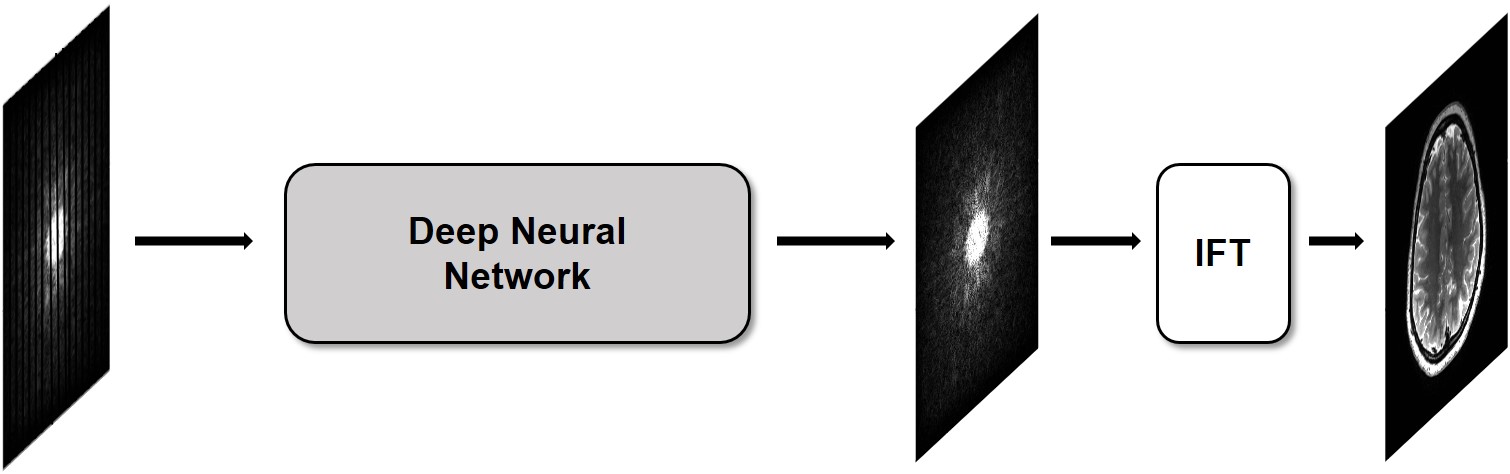}}
\vspace{-0.4cm}
\centerline{\mbox{(d)}}
\caption{Deep learning frameworks for accelerated MRI: (a) image domain learning  \cite{lee2018deep,han2017deep}, (b) cascaded network \cite{hammernik2018learning,wang2016accelerating,schlemper2018deep}, (c) AUTOMAP \cite{zhu2018image},
and (d) the proposed $k$-space learning.  IFT: Inverse Fourier transform. }
\vspace*{-0.5cm}
\label{fig:deepMR}
\end{figure}

Although the  end-to-end recovery approach like AUTOMAP \cite{zhu2018image} may directly recover the image without ever interpolating the missing $k$-space samples (see Fig.~\ref{fig:deepMR}(c)),
it works only for the sufficiently small size images 
due to its huge memory requirement for fully connected layers.
Accordingly, most of the popular deep learning MR reconstruction algorithms
are either in the form of  image domain post-processing as shown in Fig.~\ref{fig:deepMR}(a) \cite{lee2018deep,han2017deep}, or iterative updates between the $k$-space and the image domain using a cascaded network as shown in Fig.~\ref{fig:deepMR}(b) \cite{hammernik2018learning,wang2016accelerating,schlemper2018deep,aggarwal2019modl}.

One of the main purposes of this paper is  to reveal that the aforementioned approaches are not all the available options for MR deep learning, but there exists 
another  effective
deep learning approach.
In fact, as illustrated in Fig.~\ref{fig:deepMR}(d),
the proposed deep learning approach directly interpolates the missing $k$-space data so that accurate reconstruction can be obtained by simply taking
the Fourier transform of the interpolated $k$-space data.
In contrast to AUTOMAP  \cite{zhu2018image},  our network is implemented in the form of convolutional neural network (CNN) 
without requiring fully connected layer, so the GPU memory requirement for the proposed
$k$-space deep learning is minimal.

In fact,  our $k$-space deep learning scheme is inspired by the success of structured low-rank Hankel matrix approaches \cite{shin2014calibrationless,haldar2014low,jin2016general,ongie2016off,ye2016compressive},
exploiting  concise signal representation in the $k$-space domain   thanks to the duality between
the structured low-rankness in the $k$-space  and the image domain sparsity.
In addition, the recent theory of deep convolutional framelets \cite{ye2017deep} showed that
an encoder-decoder network can be regarded as a signal representation that emerges from  Hankel matrix decomposition.
Therefore,  by synergistically combining these findings, we  propose
a novel  $k$-space deep learning algorithms that  perform better and generalize well.
%
We further show that our deep learning approach for
 $k$-space interpolation can handle general $k$-space sampling patterns beyond the Cartesian trajectory, such as radial, spiral, etc. 
Moreover, our theory and empirical results  also shows that
multi-channel calibration-free $k$-space interpolation can be easily realized using the
proposed framework by simply stacking multi-coil $k$-space data along the channel direction as an input to feed in the neural
network.

We are aware of recent $k$-space neural network approaches called the scan-specific robust artificial neural networks for $k$-space interpolation (RAKI)
\cite{akccakaya2019scan}. Unlike the proposed method, RAKI considers scan-specific learning without training data, so the neural network weights needs to 
be recalculated for each $k$-space input data. On the other hand, the proposed method exploits the generalization capability of the neural network.  More specifically, even with the same trained weights, the trained neural network can generate diverse signal representation depending on the input  thanks to the combinatorial nature 
of the ReLU nonlinearity. This
 makes the neural network expressive and generalized well to the unseen test data. The theoretical origin of the expressiveness will be also discussed in this paper.

After the original version of this work was available on Arxiv,  there appear several  deep learning
algorithms exploiting $k$-space learning \cite{pramanik2018off,aggarwal2018multi}. These works are based on  hybrid formulation (see Fig.~\ref{fig:deepMR}(b))  and utilize the
deep neural network as a regularization term for $k$-space denoising. On the other hand,
 the our method
exploits the nature of  deep neural network as a generalizable and expressive $k$-space representation to directly
interpolate the missing $k$-space data.


\section{Mathematical Preliminaries}

\subsection{Notations}

In this paper, matrices are denoted by bold upper case letters, i.e. $\Ab,\Bb$, whereas
the vectors are represented by bold lower cases letters, i.e. $\xb,\yb$.
In addition, $[\Ab]_{ij}$ refers to the $(i,j)$-th element of the matrix $\Ab$, and $x[i]$ denotes the $i$-th element of the 
vector $\xb$.
The notation $\overline \vb\in \Rd^d$ for a vector $\vb\in \Rd^d$ denotes its flipped version, i.e. the indices of $\vb$ are time-reversed
such that $\overline\vb[n]=\vb[-n]$.
The $N\times N$ identity matrix is denoted as $\Ib_N$, while $\oneb_N$ is an $N$-dimensional vector
with 1's. 
The superscript $^T$ and $^\top$ for a matrix or vector denote
the transpose and Hermitian transpose, respectively.
$\Rd$ and $\Cd$ denote the real and imaginary fields, respectively.
$\Rd_+$ refers to the nonnegative real numbers.

\subsection{Forward Model for Accelerated MRI}

 The spatial Fourier transform of an arbitrary smooth function $x:\RR^2\to\RR$ is defined by 
\begin{align*}
\hat{x}(\bk)=\mathcal{F}[x](\bk):=\int_{\RR^d} e^{-\iota\bk\cdot \rb}x(\rb)d\rb,
\end{align*}
with spatial frequency $\bk\in\RR^2$ and $\iota=\sqrt{-1}$.
Let  $\{\bk_n\}_{n=1}^N$, for some integer $N\in\NN$, be a collection of finite number of sampling points of the $k$-space
 confirming to the Nyquist sampling rate. 
 Accordingly, the discretized $k$-space data  $\widehat\xb\in \Cd^N$  is introduced by
\begin{equation}\label{eq:coil}
\widehat \xb = \begin{bmatrix} \hat x[0] &\cdots & \hat x[N-1]\end{bmatrix}, 
\quad \mbox{where} \quad \hat x[i] = \widehat x(\kb_i) . 
\end{equation}
  For a given under-sampling pattern $\Lambda$ for accelerated MR acquisition, let 
  the downsampling operator $\Pc_\Lambda: \Cd^{N} \to \Cd^{N}$ 
  be defined as
  \begin{eqnarray}
  \left[\Pc_\Lambda[\hat \xb] \right]_i= \begin{cases}  \widehat x[i], &i \in \Lambda \\
0, &  \mbox{otherwise} \end{cases}   \   .
  \end{eqnarray}
Then, the under-sampled  $k$-space data is given by
\begin{eqnarray}\label{eq:fwd}
\hat \yb & :=\Pc_\Lambda[\hat \xb] 
\end{eqnarray}

\subsection{Low-Rank Hankel Matrix Approaches} 
\label{sec:theory}

From the undersampled data in \eqref{eq:fwd}, CS-MRI  \cite{lustig2007sparse,jung2009k} attempts to find the feasible solution  that has minimum non-zero support in some sparsifying
transform domain. This can be achieved
by finding a function $z:\Rd^2\to \Rd$ such that 
\begin{eqnarray}
\min_z &  \|\Tc z\|_1 \notag\\
\mbox{subject to } & \Pc_\Lambda[\hat \xb] = \Pc_\Lambda[\hat \zb]
\end{eqnarray}
where $\Tc$ denotes the image domain sparsifyting transform
and 
\begin{equation}
\widehat \zb = \begin{bmatrix} \widehat z(\kb_0) & \cdots  & \widehat z(\kb_{N-1}) \end{bmatrix}^T  \  .
\end{equation}
This optimization problem usually requires  iterative update between the $k$-space and the image domain after the discretization of  $z(\rb)$  \cite{lustig2007sparse,jung2009k}.

On the other hand, in  recent structured low-rank  matrix completion algorithms  \cite{shin2014calibrationless,haldar2014low,jin2016general,ongie2016off,ye2016compressive},
the compressed sensing problems was solved either by imposing the low-rank structured matrix penalty \cite{haldar2014low,ongie2016off}
or by converting it a direct $k$-space interpolation problem using low-rank structure matrix completion \cite{shin2014calibrationless,jin2016general}.
 More specifically, let $\hank_d(\widehat \xb)$ denote a 
 Hankel matrix constructed from the $k$-space measurement $\widehat \xb$ in \eqref{eq:coil}, where $d$ denotes the
 matrix pencil size (for more details on the construction of Hankel matrices and their relation to the convolution, see Section I in Supplementary Material).
Then,
 if the underlying signal $x(\rb)$ in the image domain is sparse and  described as the signal with the finite rate of innovations (FRI) with rate $s$ \cite{vetterli2002sampling},
  the associated Hankel matrix $\hank_d(\hat \xb)$ with  $d>s$
 is low-ranked  \cite{ye2016compressive,jin2016general,ongie2016off}.
  Therefore, if some of $k$-space data  are missing,
we can construct an appropriate weighted Hankel matrix with missing elements such that the missing elements are recovered 
using low-rank Hankel matrix completion approaches \cite{candes2009exact}:
\begin{eqnarray}\label{eq:EMaC}
(P)
 &\min\limits_{\widehat \zb\in \Cd^N } & \rank~ \hank_d (\widehat \zb)  \\
&\mbox{subject to } & \Pc_\Lambda[\widehat\xb ] = \Pc_\Lambda[\widehat \zb]  \nonumber  \  .
\end{eqnarray}
While the low-rank Hankel matrix completion problem $(P)$  can be solved in various ways, 
one of the main technical huddles is its relatively large computational complexity for matrix factorization and  large memory 
requirement for storing Hankel matrix. 
Although several new approaches have been proposed to solve these problems \cite{ongie2017fast}, the following section shows that a deep learning approach is a novel and efficient way to solve this problem.

\section{Main Contribution}

\subsection{ALOHA as a signal representation}

Consider the following image regression problem under the  low-rank Hankel matrix constraint:
\begin{eqnarray}
 \quad & \min_{\widehat\zb\in \Cd^{N}}  & \left\|x- \Fc^{-1}[ \widehat\zb]\right\|^2 \label{eq:imgcost}  \\
&\mbox{subject to }  &\rank~ \hank_{d}\left( \widehat\zb\right) =s ,  \quad
  \Pc_\Lambda \left[\widehat\xb\right]= \Pc_\Lambda \left[\widehat \zb\right],\quad  \label{eq:ccost}
\end{eqnarray}
where  $s$ denotes the estimated rank. In the above formulation,  the cost in \eqref{eq:imgcost} is defined in the image domain to minimize the
errors in the image domain, whereas
the low-rank Hankel matrix constraint in \eqref{eq:ccost} is imposed in the $k$-space after the $k$-space weighting.

Now, we convert the complex-valued constraint in \eqref{eq:ccost}
to a  real-valued constraint.
The procedure is as follows. First, 
the operator  $\Rbc:\Cd^{N } \to \Rd^{N\times 2}$ is defined as 
\begin{eqnarray}\label{eq:Rbc}
\Rbc[\widehat\zb]:= \begin{bmatrix}\re(\hat \bz) & \im (\hat \bz) \end{bmatrix},\quad \forall \widehat\zb \in \Cd^{N}
\end{eqnarray}
where $\re(\cdot)$ and $\im(\cdot)$ denote the real and imaginary part of the argument.
Similarly, we define its inverse operator  $\Rbc^{-1}:\Rd^{N\times 2} \to \Cd^{N}$ as 
\begin{eqnarray}\label{eq:Za}
\Rbc^{-1}[\widehat\Zb]:= \hat \zb_1+\iota \hat \zb_2,\quad \forall \widehat\Zb :=[\zb_1~ \zb_{2}]\in \Rd^{N\times 2}
\end{eqnarray}
Then, as shown in Section II in Supplementary Material, 
we can approximately convert \eqref{eq:ccost} to an 
optimization problem with real-valued constraint:
\begin{eqnarray}
(P_A) &\min_{\widehat\zb \in \Cd^{N}}  & \left\|x- \Fc^{-1}[\widehat\zb ]\right\|^2  
\end{eqnarray}
\begin{eqnarray*}
\mbox{subject to}& \rank  \hank_{d|2}\left( \Rbc[\widehat\zb ]\right) =Q  \leq 2s , \notag  \\
& \Pc_\Lambda \left[\widehat\xb\right]= \Pc_\Lambda \left[\widehat \zb \right] .\quad   \label{eq:rcost}
\end{eqnarray*}

In the recent theory of deep convolutional framelets \cite{ye2017deep}, 
this low-rank constraint optimization problem was addressed  using  learning-based signal representation.
More specifically,
 for any  $\widehat \zb\in \Cd^N$, let the Hankel structured matrix $\hank_{d|2}\left(\Rbc[\widehat \zb ]\right)$
 have  the singular value decomposition
$\Ub \Sigmab \Vb^{\top}$, 
where $\Ub =[\bu_1 \cdots \bu_Q] \in \Rd^{N\times Q}$ and $\Vb=[\bv_1\cdots \bv_Q]\in \Rd^{2d \times Q}$ denote the left and the right singular vector bases matrices, respectively;
$\Sigmab=(\sigma_{ij})\in\mathbb{R}^{Q\times Q}$ is the diagonal matrix with singular values.  
Now, consider
matrix pair  $\Psib$, $\tilde \Psib\in \Rd^{2d \times Q}$ 
\begin{eqnarray}\label{eq:Psi}
\Psib:=
\begin{pmatrix}
\psib^{1}_{1} & \cdots & \psib^{1}_{Q}
\\
\psib^{2}_{1} & \cdots &\psib^{2}_{Q}
\end{pmatrix}
\,\text{ and }\,
\widetilde{\Psib}:=
\begin{pmatrix}
\tilde\psib^{1}_{1} & \cdots & \tilde\psib^{1}_{Q}
\\
\tilde\psib^{2}_{1} & \cdots &\tilde\psib^{2}_{Q}
\end{pmatrix}
\end{eqnarray}
that 
satisfy the low-rank  projection constraint:
\begin{eqnarray}\label{eq:projection}
 \Psib \tilde \Psib^{\top} = \Pb_{R(\Vb)} ,
 \end{eqnarray}
 where $\Pb_{R(\Vb)}$ denotes the projection matrix to the range space of $\Vb$.
 We further introduce the generalized pooling and unpooling matrices $\Phib,\widetilde\Phib\in \Rd^{N\times M}$ \cite{ye2017deep} that
 satisfies the condition
\begin{eqnarray}\label{eq:projectionU}
 \widetilde\Phib \Phib^{\top} =\Ib_N, 
 \end{eqnarray} 
 Using Eqs. \eqref{eq:projection} and \eqref{eq:projectionU},  we can obtain the following matrix equality:
\begin{eqnarray}\label{eq:B}
\hank_{d|2}\left(\Rbc[\widehat \zb ]\right) = \widetilde\Phib \Phib^{\top}\hank_{d|2}\left(\Rbc[\widehat \zb ]\right) \Psib \tilde \Psib^{\top} =  \widetilde\Phib \Cb \tilde \Psib^{\top},  \label{eq:equiv}
\end{eqnarray}
where 
\begin{eqnarray}\label{eq:C}
\Cb :=  \Phib^{\top}\hank_{d|2}\left(\Rbc[\widehat \zb ]\right)  \Psib \quad   \in  \Rd^{N\times Q}
\end{eqnarray}
By taking the generalized inverse of Hankel matrix,  \eqref{eq:B} can be converted
to the framelet basis representation  \cite{ye2017deep}.
Moreover, one of the
most important observations in  \cite{ye2017deep}
is  that  the resulting framelet basis representation  can be equivalently represented by
single layer encoder-decoder convolution architecture:
\begin{eqnarray} \label{eq:decomp0}
\Rbc[\widehat \zb ]= 
 \left(\widetilde\Phib \Cb\right) \circledast g(\tilde \Psib),  ~\mbox{where}~
 \Cb =  \Phib^\top \left(  \Rbc[\widehat \zb ]\circledast  h(\Psib)\right) 
\label{eq:finsuf}
\end{eqnarray}
and  
 $\circledast$ denotes the multi-channel input multi-channel output convolution.
The second and the first part of \eqref{eq:finsuf} correspond to the encoder and decoder layers with the corresponding
 convolution filters  
$h(\Psib) \in\RR^{2d \times Q}$ and $g\left(\widetilde{\Psib}^{(\jmath)}\right)\in\RR^{d Q\times 2}$:
\begin{align*}
&h(\Psib):=
\begin{pmatrix}
\overline\psib^{1}_{1} & \cdots & \overline\psib^{1}_{Q}
\\
\overline\psib^{2}_{1} & \cdots & \overline\psib^{2}_{Q}
\end{pmatrix}
\ , ~~
g\left(\widetilde{\Psib}\right):=
\begin{pmatrix}
\widetilde{\psib}^{1}_{1} & \widetilde{\psib}^{2}_{1}
\\
\vdots & \vdots
\\
\widetilde{\psib}^{1}_{Q} &  \widetilde{\psib}^{2}_{Q}
\end{pmatrix},
\end{align*}
which are obtained by reordering the matrices $\Psib$ and $\widetilde\Psib$ in \eqref{eq:Psi}.
Specifically,  $\overline \psib_i^{1}\in \Rd^{d}$ (resp. $\overline \psib_i^{2}\in \Rd^d$) denotes the $d$-tap
encoder convolutional filter applied to the  real (resp. imaginary) component of the $k$-space data to 
generate the $i$-th channel output.
In addition, $g(\tilde\Psib)$ is a reordered version of $\tilde\Psib$ so that 
 $\tilde \psib_i^{1}\in \Rd^d$ (resp. $\tilde \psib_i^{2}\in \Rd^d$) corresponds to the $d$-tap
decoder convolutional filter to generate the  real (resp. imaginary) component of the  $k$-space data by
convolving with  the $i$-th channel input.
We can further use recursive application of
encoder-decoder representation for the resulting framelet coefficients $\Cb$ in \eqref{eq:decomp0}.
In Corollary 4 of our companion paper \cite{ye2019cnn}, we showed that 
the recursive application of the encoder-decoder
operations across the layers
increases the net length of the convolutional filters. 
%

%

Since
\eqref{eq:finsuf} is a general form of the signals that are associated with a Hankel structured matrix,
 we are interested in using it to  estimate bases for $k$-space interpolation.
 Specifically, we consider a complex-valued signal space $\Hbc$ determined by the filters $\Psib$ and $\tilde \Psib$: 
\begin{eqnarray}\label{eq:H0}
\Hbc{(\Psib,\tilde\Psib)} &=& \left\{  \zb  \in \Cd^{N} \,\Big|\,\  \Rbc[ \zb ]=  \Phib^\top \left(\Cb \circledast g(\tilde \Psib)\right), \right. \notag \\
&& \left.  \Cb = (\tilde\Phib  \Rbc[ \zb]) \circledast   h(\Psib) \right\} \  . 
\end{eqnarray}
Then, the ALOHA formulation $P_A$ can be equivalently represented by
\begin{eqnarray}
(P_A') & \min\limits_{ \widehat\zb  \in \Hbc(\Psib,\tilde\Psib)}\min\limits_{ \Psib,\tilde\Psib}  &\left\|x- \Fc^{-1}[ \widehat\zb ]\right\|^2 \notag  \\
&\mbox{subject to } &
 \Pc_\Lambda \left[\widehat\xb\right]= \Pc_\Lambda \left[\widehat \zb \right],\quad  \notag
\end{eqnarray}
In other words, ALOHA is  to find the optimal filter $\Psib,\tilde\Psib$ and the associated $k$-space
signal $\widehat\zb \in \Hbc(\Psib,\tilde\Psib)$  that satisfies the data consistency conditions.
In contrast to the standard CS approaches in which signal presentation in the image domain is separately applied
from the data-consistency constraint in  the $k$-space,
 the success of ALOHA over CS can be contributed to more efficient signal representation in the $k$-space domain
 that simultaneously take care of the data consistency in the $k$-space domain.

\subsection{Generalization and Depth}
 

To allow  training for neural networks, the problem formulation
in $(P_A')$  should be decoupled into two steps: the learning phase to estimate $\Psib,\tilde\Psib$
from the training data,  and the inference phase to estimate the
interpolate signal $\widehat \zb$ for the given filter set $\Psib,\tilde\Psib$.
Although we have revealed the relations between ALOHA  and encoder-decoder architecture, 
the derivation is for specific input signal and it is not clear how 
 the relations would translate when training is performed over multiple training data set,
and the trained network can be  generalized to the unseen test data.
Given that
the  sparsity prior in dictionary learning enables the selection of appropriate basis functions from the dictionary for each given
input, one may conjecture that
there should be similar mechanisms in deep neural networks that enable adaptation  to the specific input signals.

\begin{figure}[!h] 	
\centering
{\includegraphics[width=0.7\linewidth]{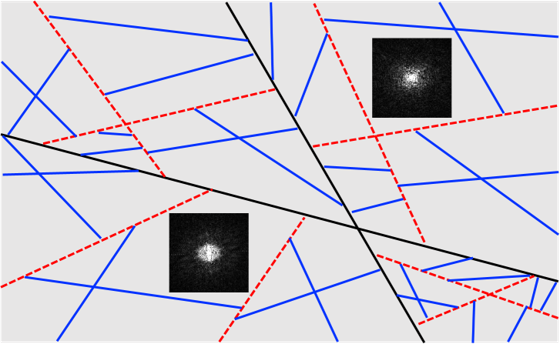}}
\caption{An example of $\Rd^2$ input space partitioning for the case of two-channel three-layer ReLU neural network.
Depending on input $k$-space data, a partition and its associated linear representation are selected. }
\label{fig:input}
\end{figure}

In Section IV of Supplementary Material,  we show that the ReLU nonlinearities indeed plays a critical role
in  the adaptation and generalization.
In fact, 
in our companion paper \cite{ye2019cnn},  
we have shown that ReLU offers  combinatorial convolution frame basis selection depending on each input image.
More specifically,
 thanks to ReLU, a trained filter set produce  large number of 
  partitions in the input space as shown in Fig.~\ref{fig:input},
 in which each region shares the same linear signal representation. 
Therefore, depending on each $k$-space  input data,  a particular region and its associated linear representation are selected.
Moreover,  
we show that
the number of input space partition and the
associated linear representation increases exponentially with the depth, channel and the skipped connection. 
By synergistically exploiting the efficient signal representation  in the $k$-space domain,
this enormous expressivity from the same filter sets can make the $k$-space deep neural
network more powerful than the conventional image domain learning.


%
For the more details on the theoretical aspect of  deep neural networks, see  Section IV of Supplementary Material
or our companion paper \cite{ye2019cnn}.

%
%
%

\begin{figure}[!t] 	
\centerline{\includegraphics[width=1\linewidth]{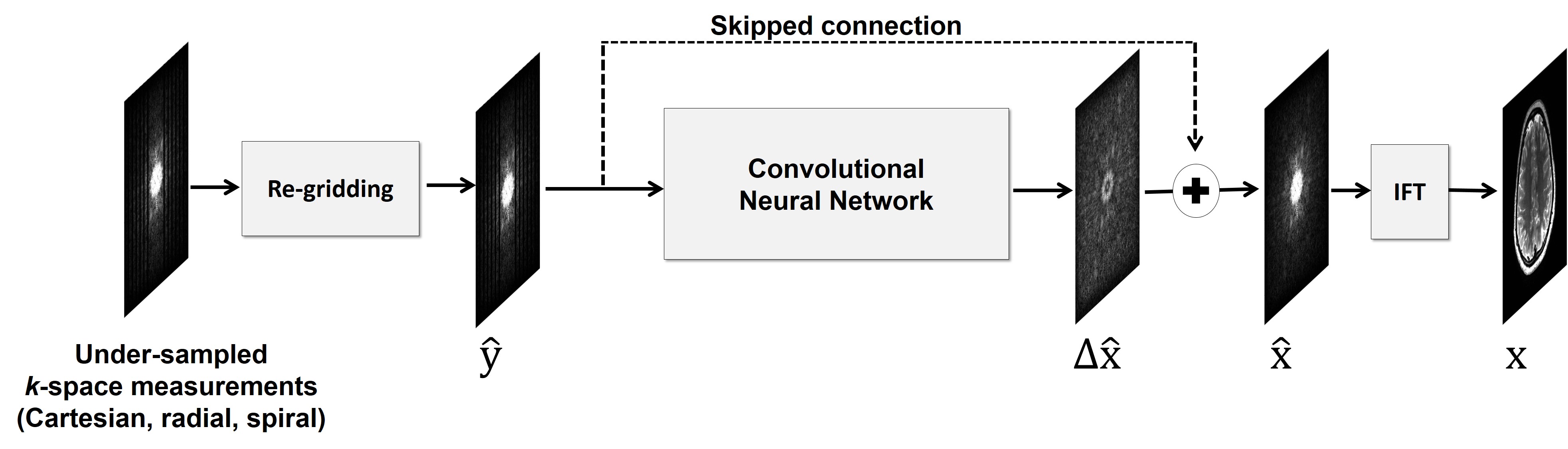}}
\vspace*{-0.7cm}
\centerline{\mbox{(a)}}
\vspace*{0.2cm}
\centerline{\includegraphics[width=1\linewidth]{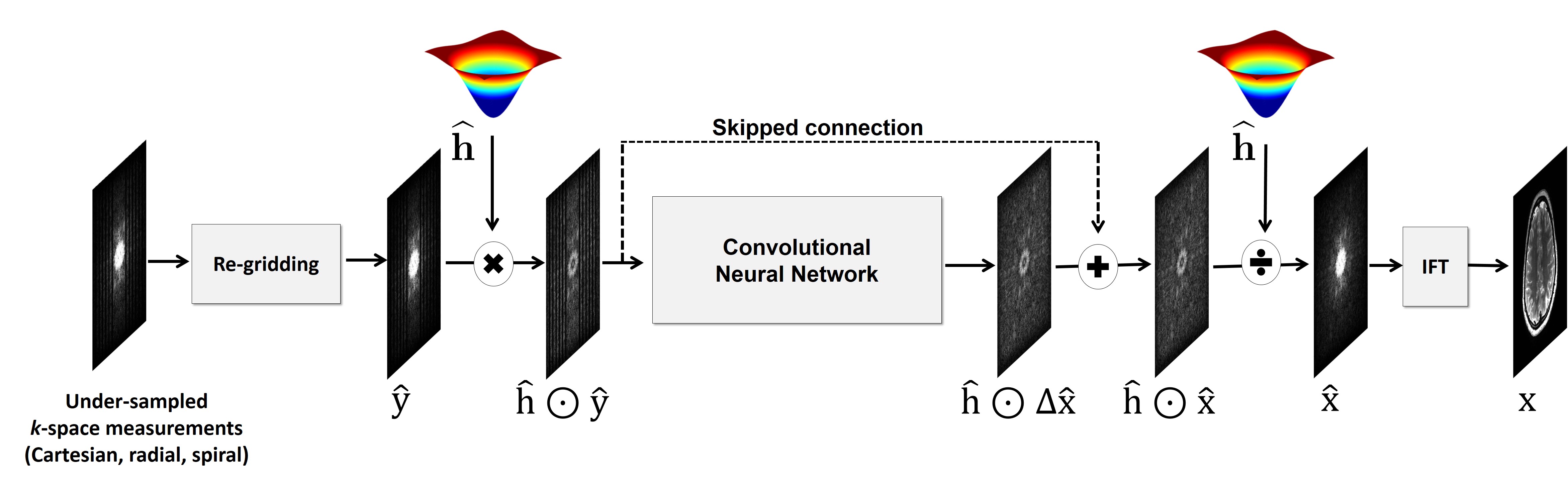}}
\vspace*{-0.7cm}
\centerline{\mbox{(b)}}
\caption{Overall reconstruction flows of the proposed method with (a) skipped connection, and (b) skipped connection and weighting layer. IFT denotes the inverse Fourier transform. For the parallel imaging, the input and output are multi-coil $k$-space data, after stacking them along
the channel direction. }
\label{fig:architecture}
\end{figure}

\subsection{Extension to parallel imaging}

In \cite{jin2016general}, we formally showed that when
$\{\hat \xb_i\}_{i=1}^P$ denote the $k$-space measurements from $P$ receiver coils,
the following extended Hankel structured matrix is low-ranked:
\begin{eqnarray}\label{eq:PHankel}
\hank_{d|P}(\widehat\Xb) = \begin{bmatrix} \hank_d(\widehat\xb_1) & \cdots & \hank_d(\widehat\xb_P) \end{bmatrix}
\end{eqnarray}
where $$\widehat\Xb=\begin{bmatrix} \widehat\xb_1 & \cdots & \widehat\xb_P \end{bmatrix} \in \Cd^{N\times P} \ .$$
Thus, similar to the single channel cases, 
  the date-driven decomposition of the extended Hankel 
matrix in \eqref{eq:PHankel} can be represented by stacking the each $k$-space data along the channel
direction and applies the deep neural network for the given multi-channel data.
Therefore, except the number of input and output channels,  the network
structure for parallel imaging data is  identical to the single channel $k$-space learning. 


\subsection{Sparsification}

To further improve the performance of the structured matrix completion approach, in \cite{ye2016compressive}, we showed that 
even if the image $x(\rb)$ may not be sparse,   it can be often converted to a sparse signal.

 For example, the outmost skipped connection for the residual learning is another way to make the signal sparse.
Note that  fully sampled $k$-space data $\hat \xb$ can be represented by
$$\widehat \xb = \widehat\yb + \Delta \widehat\xb,$$
where $\widehat\yb$ is the undersampled $k$-space measurement in \eqref{eq:fwd},
and $\Delta\widehat \xb$ is the residual part of $k$-space data that should be estimated.
In practice, some of the low-frequency part of $k$-space data including the DC component
are acquired in the undersampled measurement so that the image component from the residual
$k$-space data $\Delta\widehat\xb$ are mostly high frequency signals, which are sparse. Therefore,
$\Delta\widehat\xb$ has low-rank Hankel matrix structure, which can be effectively processed using the deep neural network.
This can be easily implemented using a skipped connection  before the deep neural network as shown in Fig.~\ref{fig:architecture}(a).
However, the skipped connection also works beyond the sparsification.  
In our companion paper \cite{ye2019cnn} (which is also repeated in Section IV of Supplementary Material), we showed that
the skipped connection at the inner layers makes the  frame basis more expressive.
Therefore,  we conjecture that the skipped connections play dual roles in our $k$-space learning.

Second, we can  convert a signal to an innovation signal  using a 
 shift-invariant transform represented by the whitening filter $h$ such that
 the resulting innovation signal
  $z = h \ast x$
 becomes an FRI signal \cite{vetterli2002sampling}.  
  For example,  many MR images can be sparsified using finite difference or wavelet transform
 \cite{jin2016general}.  
 This implies that the Hankel matrix from the weighted $k$-space data,
 $\hat z (\kb) = \hat h(\kb)\hat x (\kb)$ are low-ranked, 
 where the weight $\hat h(\kb)$ is determined
  from the finite difference or Haar wavelet transform \cite{jin2016general,ye2016compressive}.
  Thus, the deep neural network is applied to the weighted $k$-space data to estimate
  the missing  spectral data $\hat  h(\xb)\hat x(\kb)$, after which the original $k$-space data is obtained
  by dividing with the same weight, i.e. $\hat x (\kb) =  \hat z(\kb)/\hat h(\kb)$.
 This can be easily implemented using a weighting and unweighting layer as shown in Fig.~\ref{fig:architecture}(b).

In this paper, we consider these
two strategies to investigate which strategy is better for different sampling trajectories.

\subsection{Extension to General Sampling Patterns}

Since the Hankel matrix formulation in ALOHA  is based on the Cartesian coordinate, we add extra regridding layers to handle the non-Cartesian sampling trajectories. 
Specifically, for radial and spiral trajectories, the non-uniform fast Fourier transform (NUFFT) was used to perform the regridding to the Cartesian coordinate. 
For Cartesian sampling trajectory, the regridding layer using NUFFT is  not necessary, and
we instead perform a zero-filling in the unacquired $k$-space regions as an initialization step.

\section{Implementation}

\begin{figure}[!t] 	
\centering
{\includegraphics[width=1\linewidth]{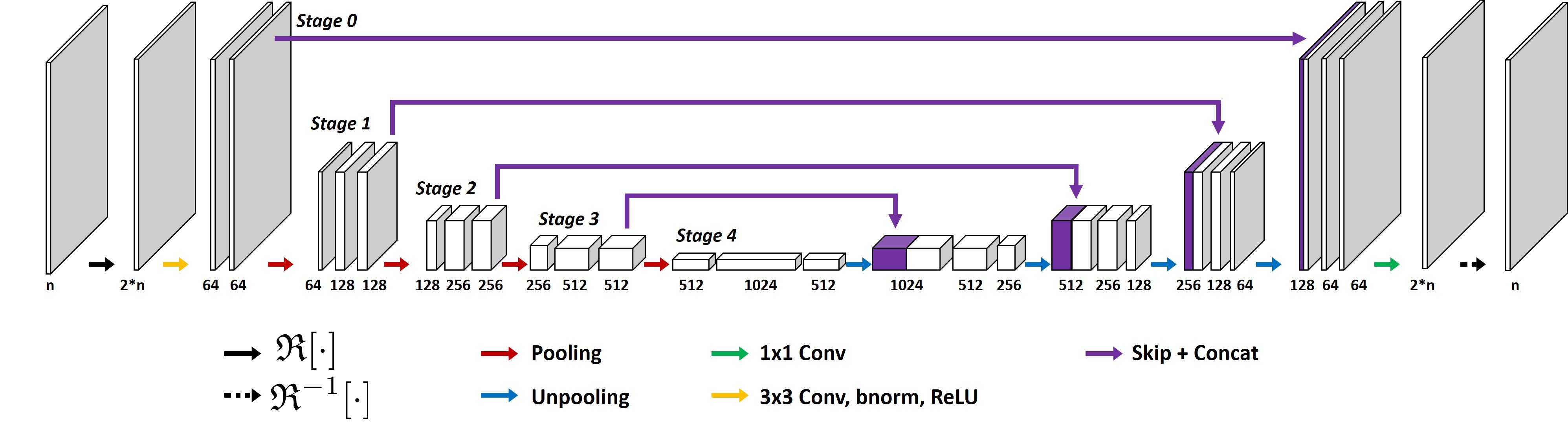}}
\caption{A network backbone of the proposed method. The input and output are complex-valued.}
\label{fig:backbone}
\end{figure}

\subsection{Network Backbone}

The network backbone  follows the  U-Net\cite{ronneberger2015u} which consists of convolution, batch normalization, rectified linear unit (ReLU), and contracting path connection with concatenation as shown in Fig. \ref{fig:backbone}.
Here, the input and output are the complex-valued $k$-space data, while $\Rbc[\cdot]$ and $\Rbc^{-1}[\cdot]$ denote the operators 
in \eqref{eq:Rbc} and \eqref{eq:Za}, respectively, that convert complex valued input to two-channel real value signals and vice versa.
For parallel imaging, 
multi-coil $k$-space data are given as input and output after stacking them along channel direction. Specially,  in our parallel
imaging experiments, we use eight coils $k$-space data.

The yellow arrow  is the basic operator that consists of $3 \times 3$ convolutions followed by a rectified linear unit (ReLU) and batch normalization.
The same operation exists  between the separate blocks at every stage, but
the yellow arrows are omitted for visibility.
A red and blue arrows are $2 \times 2$ average pooling and average unpooling operators, respectively, located between the stages.
A violet arrow is the skip and concatenation operator. 
A green arrow is the simple $1 \times 1$ convolution operator generating interpolated $k$-space data from multichannel data.

\subsection{Network Training}

We use the $l_2$ loss  in the image domain  for training. For this, 
the Fourier transform operator is placed as the last layer to convert the interpolated $k$-space data to
the complex-valued image domain so that  the loss values are calculated for the reconstructed image.
Stochastic gradient descent (SGD) optimizer was used to train the network. For the IFT layer, the adjoint operation from SGD is also Fourier transform.
The size of mini batch is 4, and the number of epochs in single and multi coil networks is 1000 and 500, respectively. The initial learning rate is $10^{-5}$, which gradually dropped to $10^{-6}$ until 300-th epochs. The regularization parameter was $\lambda = 10^{-4}$. 

The labels for the network were the images generated from direct Fourier inversion from
fully sampled $k$-space data. The input data for the network was the regridded down-sampled $k$-space data from Cartesian, radial, and spiral trajectories.
The details of the downsampling procedure will be discussed later.
 For each trajectory, we train the network separately.

The proposed network was implemented using MatConvNet toolbox in MATLAB R2015a environment \cite{vedaldi2015matconvnet}. 
Processing units used in this research are Intel Core i7-7700 central processing unit and GTX 1080-Ti graphics processing unit. Training time lasted about 5 days.

\section{Material and Methods}

\subsection{Data Acquisition}

The evaluations were performed on single coil and multi coils $k$-space data for various $k$-space trajectories
 such as Cartesian, radial, and spiral cases. 

\begin{figure}[!t] 	
\centering
{\includegraphics[width=0.9\linewidth]{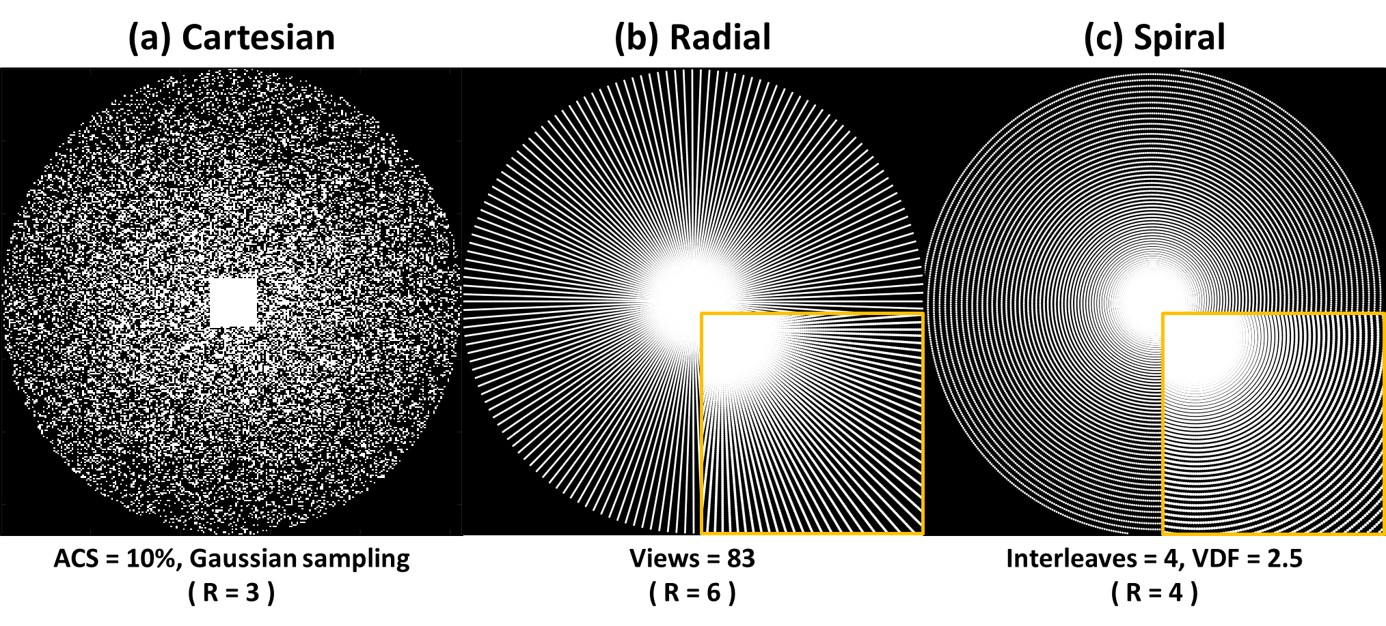}}
\caption{Various under-sampling patterns: (a) Cartesian undersampling at $R=3$, (b) radial undersampling at $R=6$, and (c) spiral undersampling at $R=4$.
Magnified views are provided for radial and spiral trajectories. }
\label{fig:trajectories}
\end{figure}

For the Cartesian trajectory, knee $k$-space dataset (http://mridata.org/) were used. The raw data were acquired from 3D fast-spin-echo (FSE) sequence with proton density weighting included fat saturation comparison by a 3.0T whole body MR system (Discovery MR 750, DV22.0, GE Healthcare, Milwaukee, WI, USA). The repetition time (TR) and echo time (TE) were 1550 ms and 25 ms, respectively. There were 320 slices in total, and the thickness of each slice was 0.6 mm. The field of view (FOV) defined $160 \times 128$ mm$^2$ and the size of acquisition matrix is $320 \times 256$. The voxel size was 0.5 mm. The number of coils is 8. Eight coils $k$-space data was used for multi-coil $k$-space deep learning. 
In addition, to evaluate the performance of the algorithm  for the single coil experiment,  
coil compression (http://mrsrl.stanford.edu/\~tao/software.html) was applied to obtain a single coil $k$-space data. 
For the Cartesian trajectory as shown in Fig. \ref{fig:trajectories}(a), the input $k$-space was downsampled to a Gaussian pattern using x4 acceleration factor in addition to the 10$\%$ auto-calibration signal (ACS) line. Therefore, the net acceleration factor is about 3 ($R=3$). 
Among the 20  cases of knee data,  18 cases were used for training, 1 case for validation, and the other for test.

For radial and spiral sampling patterns, a synthesized $k$-space data from Human Connectome Project (HCP) MR dataset (https://db.humanconnectome.org) were used.
Specifically, the multi-coil radial and spiral $k$-space data are generated using MRI simulator (http://bigwww.epfl.ch/algorithms/mri-reconstruction/).
The T2 weighted brain images contained within the HCP were acquired Siemens 3T MR system using a 3D spin-echo sequence. The TR and TE were 3200 ms and 565 ms, respectively. The number of coils was 8, but the final reconstruction was obtained as the absolute of the sum. 
The FOV was $224 \times 224 ~\rm{mm}^2$, and the size of acquisition matrix was $320 \times 320$. The voxel size was 0.7 mm. 
The total of 199 subject datasets was used in this paper. Among the 199 subject, 180 were used for network training,  10 subject  for validation, and the other subject  for test.
 Fig. \ref{fig:trajectories}(b) shows the down-sampled $k$-space radial sampling patterns. 
 The  downsampled radial $k$-space consists of only 83 spokes, which corresponds to 
$R=6$ acceleration factor compared to the 503 spokes for the fully sampled data that were used as the ground-truth.
On the other hand,  Fig. \ref{fig:trajectories}(c) shows the down-sampled spiral sampling pattern, composed of
4 interleaves that corresponds to $R=4$ acceleration compared to the
he full spiral trajectory with 16 interleaves. The spiral $k$-space trajectory was obtained with a variable density factor (VDF) of 2.5.

\begin{figure}[!b] 	
\centerline{\includegraphics[width=1\linewidth]{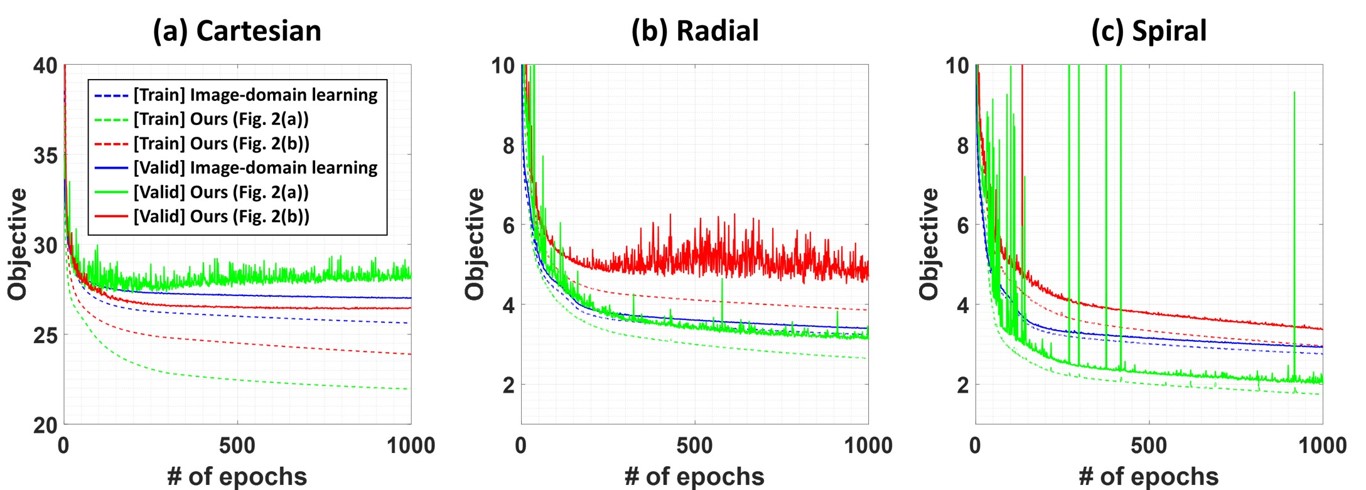}}
\caption{Objective functions of a single coil for (a) Cartesian, (b) radial, and (c) spiral trajectories. Dashed and solid lines indicate an objective function of the train and validation phase, respectively.}
\label{fig:objective}
\end{figure}

\subsection{Performance Evaluation}

For quantitative evaluation, the normalized mean square error (NMSE) value was used, which is defined as
\begin{eqnarray}
	NMSE = \frac{\sum_{i=1}^{M} \sum_{j=1}^{N} [x^*(i,j) - {x}(i, j)]^2}{\sum_{i=1}^{M}\sum_{j=1}^{N}[x^*(i,j)]^2},
\end{eqnarray}
where $x$ and $x^*$ denote the reconstructed images and ground truth, respectively. $M$ and $N$ are the number of pixel for row and column.
We also use the peak signal to noise ratio (PSNR), which is defined by
\begin{eqnarray}
	PSNR 
		 &=& 20 \cdot \log_{10} \left(\frac{NM\|x^*\|_\infty}{\| x- x^*\|_2}\right) \  .
\label{eq:psnr}		 
\end{eqnarray}
We also use the structural similarity (SSIM) index  \cite{wang2004image}, defined as
\begin{equation}
	SSIM = \dfrac{(2\mu_{x}\mu_{x^*}+c_1)(2\sigma_{x x^*}+c_2)}{(\mu_{x}^2+\mu_{x^*}^2+c_1)(\sigma_{x}^2+\sigma_{x^*}^2+c_2)},
\end{equation}
where $\mu_{x}$ is a average of $x$, $\sigma_{x}^2$ is a variance of $x$ and $\sigma_{x x^*}$ is a covariance of $x$ and $x^*$. 
There are two variables to stabilize the division such as $c_1=(k_1L)^2$ and $c_2=(k_2L)^2$.
$L$ is a dynamic range of the pixel intensities. $k_1$ and $k_2$ are constants by default $k_1=0.01$ and $k_2=0.03$. 

For extensive comparative study, we also compared with the following algorithms:
 total variation (TV) penalized CS,  ALOHA \cite{jin2016general}, and four types of CNN models including the
 variational model \cite{hammernik2018learning}, a cascade model \cite{schlemper2018deep}, the cross-domain model called KIKI network  \cite{eo2018kiki}, and an image-domain model \cite{han2017deep}. In particular,  \cite{han2017deep}  is a representative
 example of  Fig. \ref{fig:deepMR}(a).
Specifically,  the image domain residual learning using the standard U-Net backbone in Fig.~\ref{fig:backbone} was used. 
Unlike the proposed network, the input and output are an artifact corrupted image and artifact-only image, respectively \cite{jin2017deep}.
In addition,  the variational model \cite{hammernik2018learning} and the cascade model \cite{schlemper2018deep}  represent Fig. \ref{fig:deepMR}(b). The cross-domain model is formed by linking the $k$-space model in Fig. \ref{fig:deepMR}(d) and the image-domain model in Fig. \ref{fig:deepMR}(a).  
 Unfortunately, Fig. \ref{fig:deepMR}(c) does not scale well due to the enormous memory requirement, so cannot be used in the comparative study. 
 For fair comparison, the cascade \cite{schlemper2018deep} and cross-domain \cite{eo2018kiki} networks were modified for
 parallel imaging.   All the neural networks were trained using exactly the same data set.
 For  ALOHA \cite{jin2016general} and the proposed method in Fig. \ref{fig:architecture}(b),
 the  total variation based $k$-space weighting was used.


\begin{figure}[!ht] 	
\centerline{\includegraphics[width=0.9\linewidth]{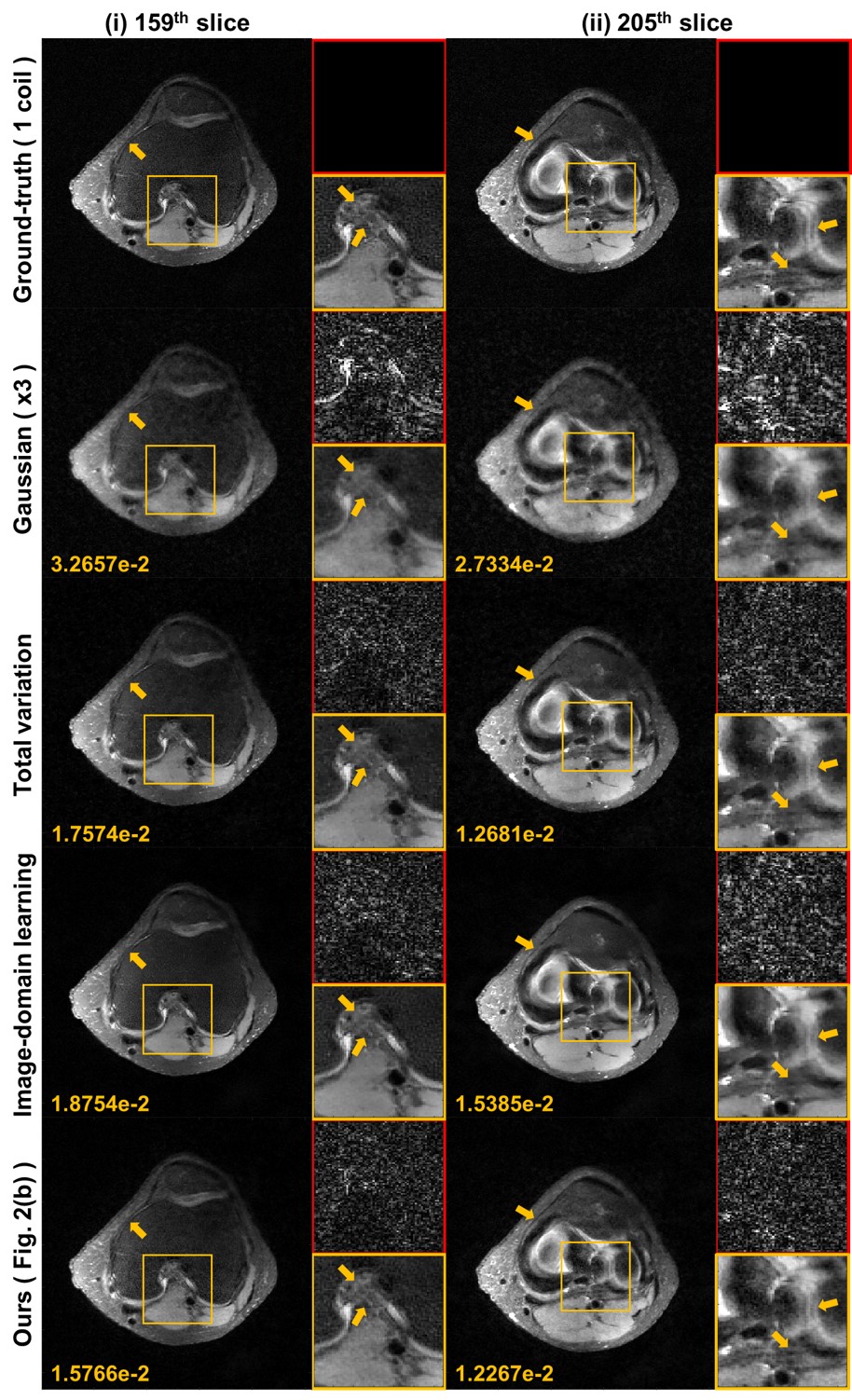}}
\caption{Reconstruction results from Cartesian trajectory at $R=3$ in single coil.
The difference images were amplified five times. Yellow and red boxes illustrate the enlarged and difference views, respectively. The number written to the images is the NMSE value.}
\label{fig:result_cartesian_axial_single}
\end{figure}

 \begin{figure}[!hbt] 	
\centerline
{\includegraphics[width=0.9\linewidth]{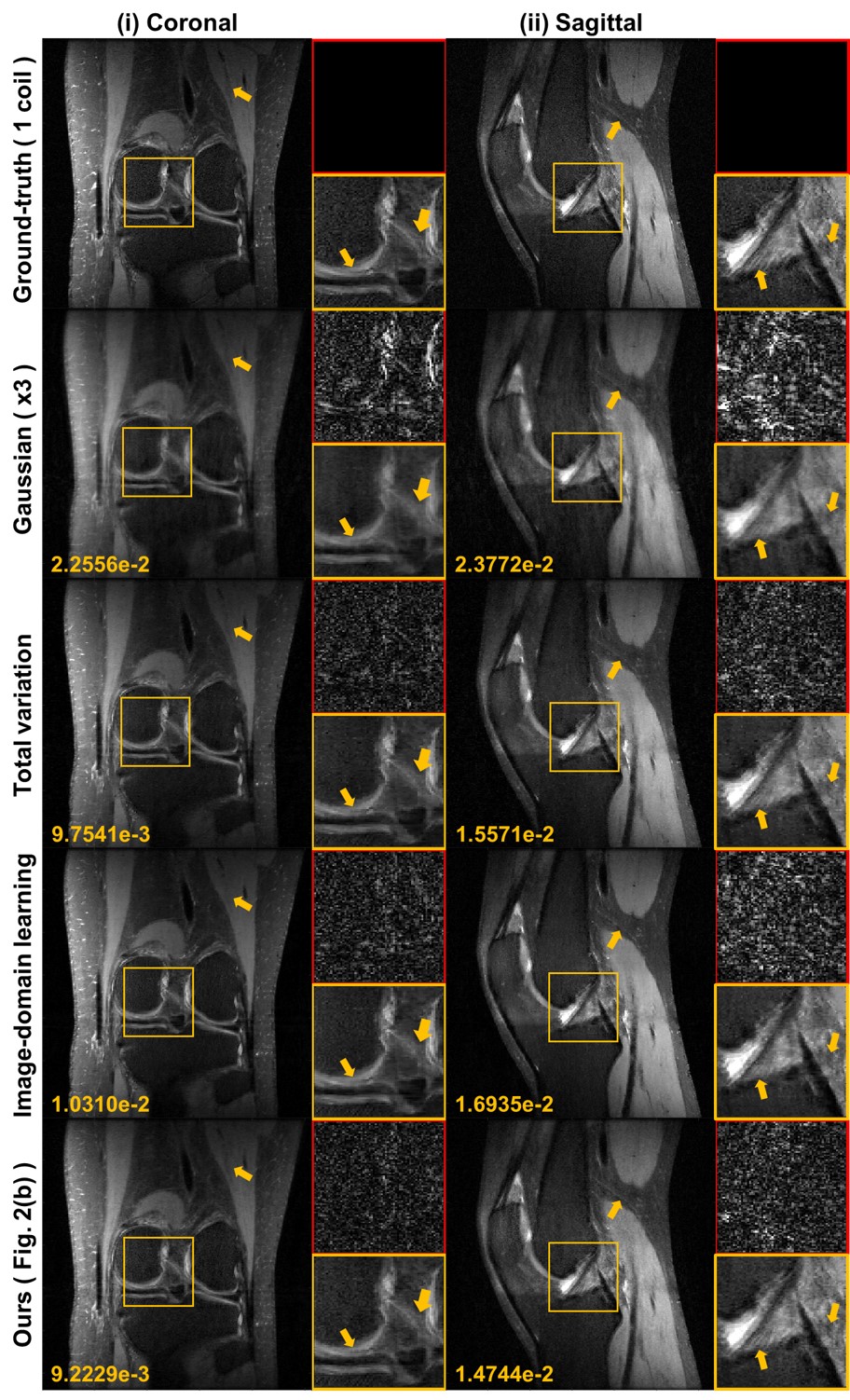}}
\caption{(i) Coronal and (ii) sagittal reformated reconstruction results from Cartesian trajectory at $R=3$ in single coil. The results from first to last rows indicate ground-truth, downsampled, total variation, image-domain learning and the proposed method. Yellow and red boxes illustrate the enlarged and difference views, respectively. The difference images were amplified five times. The number written to the images is the NMSE value. }
\label{fig:result_cartesian_single}
\end{figure}

 \section{Results}\label{sec:result}

To evaluate the performance of sparsifications  in the single coil, the proposed method was trained using the sparsifications as shown in as Figs. \ref{fig:architecture}(a)(b). Fig. \ref{fig:objective} shows the objective functions along the trajectories such as (a) Cartesian, (b) radial, and (c) spiral trajectory. In the Cartesian trajectory, the proposed network in Fig. \ref{fig:architecture}(b) produces the lowest curve in the validation phase (see red curves in Fig. \ref{fig:objective}(a)). The proposed network in Fig \ref{fig:architecture}(a) shows the best convergence during
the training, but the generalization at the test phase was not good (see green curves in Fig. \ref{fig:objective}(a)). In the non-Cartesian trajectories, the best convergence appears using the proposed network with only skipped connection in Fig. \ref{fig:architecture}(a) (see green curves in Fig. \ref{fig:objective}(b)(c)). 
Based on these convergence behaviour and generalization, the proposed network was trained with different sparsification schemes. The network in Fig. \ref{fig:architecture}(b) was trained for the Cartesian trajectory and the network in Fig. \ref{fig:architecture}(a) was used for the non-Cartesian trajectories such as radial and spiral trajectories.

\begin{table}[!t]
\centering
\resizebox{0.48\textwidth}{!}{
	\begin{tabular}{c|c|c|c|c}
		\hline
			\ \multirow{2}{*}{Metric}	& Input  			& \multirow{2}{*}{Total variation}	& Image-domain 	& Ours \\ 
			\ 							& ( 1 coil, $\times 3$ )	& 		& learning		& ( Fig. \ref{fig:architecture}(b) )	\\ \hline\hline
			\ PSNR [dB]					& 33.9058			& 35.7112			& 35.5419		& \textbf{35.9586}	\\
			\ NMSE ($\times 10^{-2}$)	& 2.7779			& 1.9067			& 1.9567		& \textbf{1.7921}	\\			
			\ SSIM						& 0.7338 			& 0.7548			& 0.7447		& \textbf{0.7664}	\\ \hline
			\ Times (sec/slice)			& - 				& 0.1344			& 0.0272		& 0.0377			\\
		\hline
	\end{tabular}
		}
\caption{Quantitative comparison from Cartesian trajectory at $R=3$ in single coil. }
\label{tbl:result_cartesian_single}
\end{table}

\begin{figure*}[!t] 	
\centerline{\includegraphics[width=0.9\linewidth]{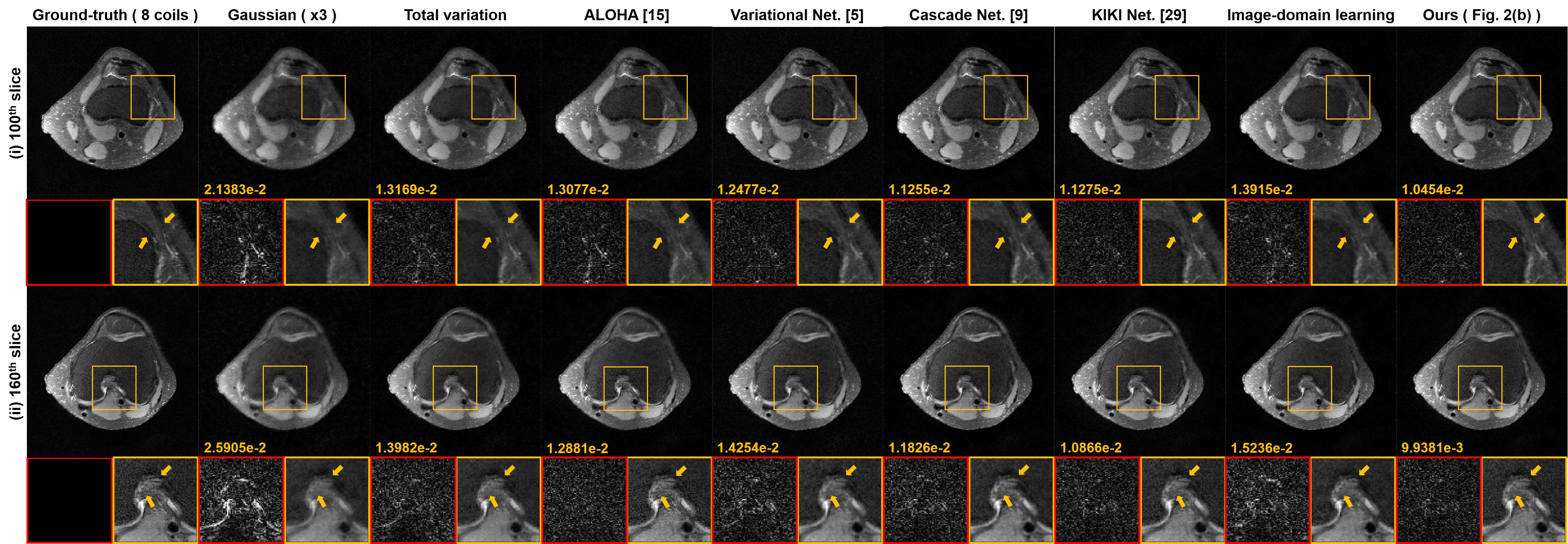}}
\centerline{\mbox{(a)}}
\vspace*{0.2cm}
\centerline{\includegraphics[width=0.9\linewidth]{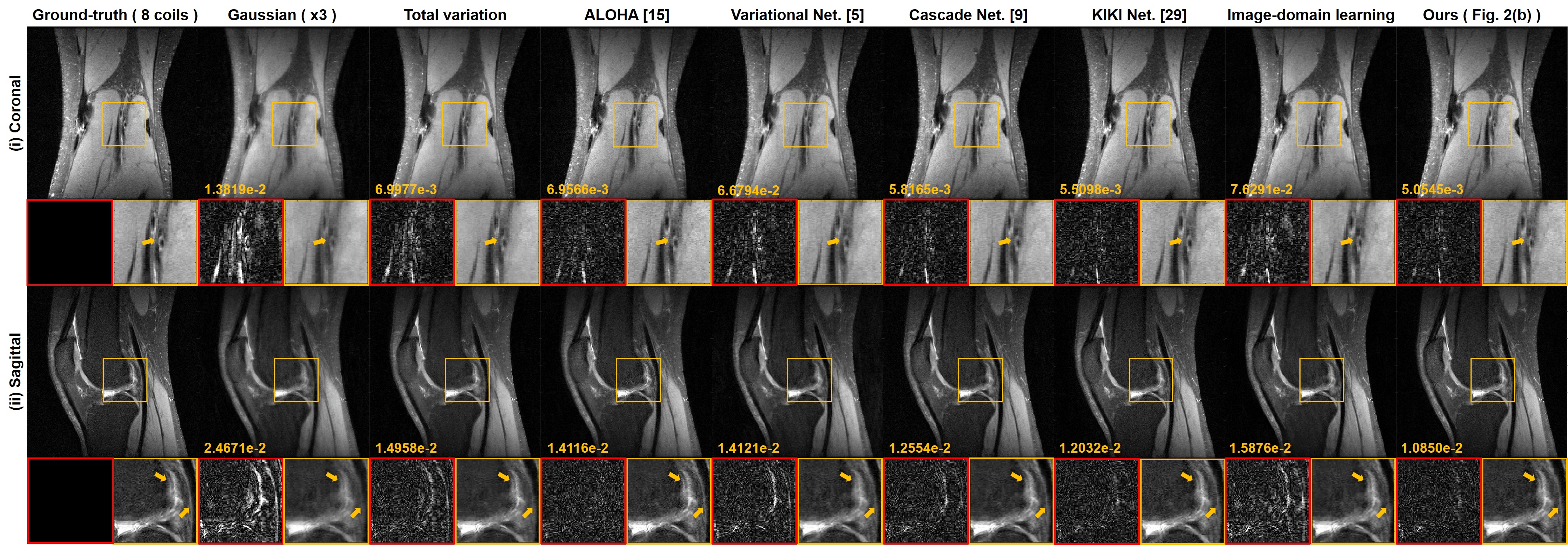}}
\centerline{\mbox{(b)}}
\caption{Reconstruction results from Cartesian trajectory at $R=3$ in multi coils: (a) axial, (b-i) coronal and (b-ii) sagittal reconstruction results. Yellow and red boxes illustrate the enlarged and difference views, respectively. The difference images were amplified five times. The number written to the images is the NMSE value.}
\label{fig:result_cartesian_multi}
\end{figure*}

\begin{table*}[!t]
\centering
\resizebox{0.95\textwidth}{!}{
	\begin{tabular}{c|c|c|c|c|c|c|c|c}
		\hline
			\ \multirow{2}{*}{Metric}	& Input	& \multirow{2}{*}{Total variation}	& ALOHA		& Variational Net.	& Cascade Net. 	& KIKI Net. & Image-domain 	& Ours \\ 
			\ 							& ( 8 coils, $\times 3$ )	& 	& \cite{jin2016general}	& \cite{hammernik2018learning}	& \cite{schlemper2018deep} & \cite{eo2018kiki}	& learning		& ( Fig. \ref{fig:architecture}(b) )	\\ \hline\hline
			\ PSNR [dB]					& 34.2021	& 36.0689	& 36.1013	& 36.1428	& 36.6055	& 36.6847	& 35.8497	& \textbf{36.9931}	\\
			\ NMSE ($\times 10^{-2}$)	& 2.3384	& 1.5826	& 1.5812	& 1.5379	& 1.4196	& 1.4086	& 1.6497	& \textbf{1.3154}	\\			
			\ SSIM						& 0.7710 	& 0.7878	& 0.7914	& 0.7966	& 0.7982	& 0.8018	& 0.7776	& \textbf{0.8087}	\\ \hline
			\ Times (sec/slice)			& - 		& 1.0938	& 16.5554	& 0.1688	& 0.2219	& 0.5438	& 0.1188	& 0.1438			\\
		\hline
	\end{tabular}
		}
\caption{Quantitative comparison from Cartesian trajectory at $R=3$ in 8 coils parallel imaging. }
\label{tbl:result_cartesian_multi}
\end{table*}

 Fig. \ref{fig:result_cartesian_axial_single} shows the results of single coil reconstruction from Cartesian trajectory using the architecture with skipped connection and weighting layer as shown in Fig.~\ref{fig:architecture}(a).
 While all the algorithms provide good reconstruction results, the proposed method most accurately
recovered  high frequency edges and textures 
as shown in the enlarged images and difference images of Fig.~\ref{fig:result_cartesian_axial_single}.
 Fig. \ref{fig:result_cartesian_single} shows the reformed images along the (i) coronal and (ii) sagittal directions. 
Again, the reformatted coronal and sagittal images by the proposed method
preserved the most detailed structures of underlying images without any artifact along the slice direction. 
 The quantitative comparison in Table~\ref{tbl:result_cartesian_single} in terms of   average PSNR, NMSE, and SSIM values also confirmed that
the proposed $k$-space interpolation method produced the best quantitative values in all area. The computation time of the
proposed method is slightly slower than the image-domain learning because of the weighting and Fourier transform operations,
but it is still about 3.5 times faster than the total variation penalized compressed sensing (CS) approach.

 Fig. \ref{fig:result_cartesian_multi} shows the parallel imaging results from  eight coil measurement. 
Because the ALOHA \cite{jin2016general} and the proposed method directly interpolate missing $k$-space, these methods clearly preserve textures detail structures as shown in  Fig. \ref{fig:result_cartesian_multi}. However, ALOHA \cite{jin2016general} is more than 100 times slower than the $k$-space deep learning as shown in Table \ref{tbl:result_cartesian_multi}.  
All CNN methods except the imaging-domain model outperform than CS methods. 
Although the cascade model \cite{schlemper2018deep} was performed with data consistency step, the method did not completely
overcome the limitations of image-domain learning. In the cross-domain learning \cite{eo2018kiki}, they proposed to train $k$-space and image-domain sequentially. 
The cross-domain network consists of deeper layers (100 layers; 25 layers $\times$ 4 individual models) than the proposed method, 
but the performance was worse than our method. As shown Table \ref{tbl:result_cartesian_multi},
the proposed method shows best performance in terms of average PSNR, NMSE, and SSIM values.

\begin{figure}[!t] 	
\centering
\centerline{\includegraphics[width=0.9\linewidth]{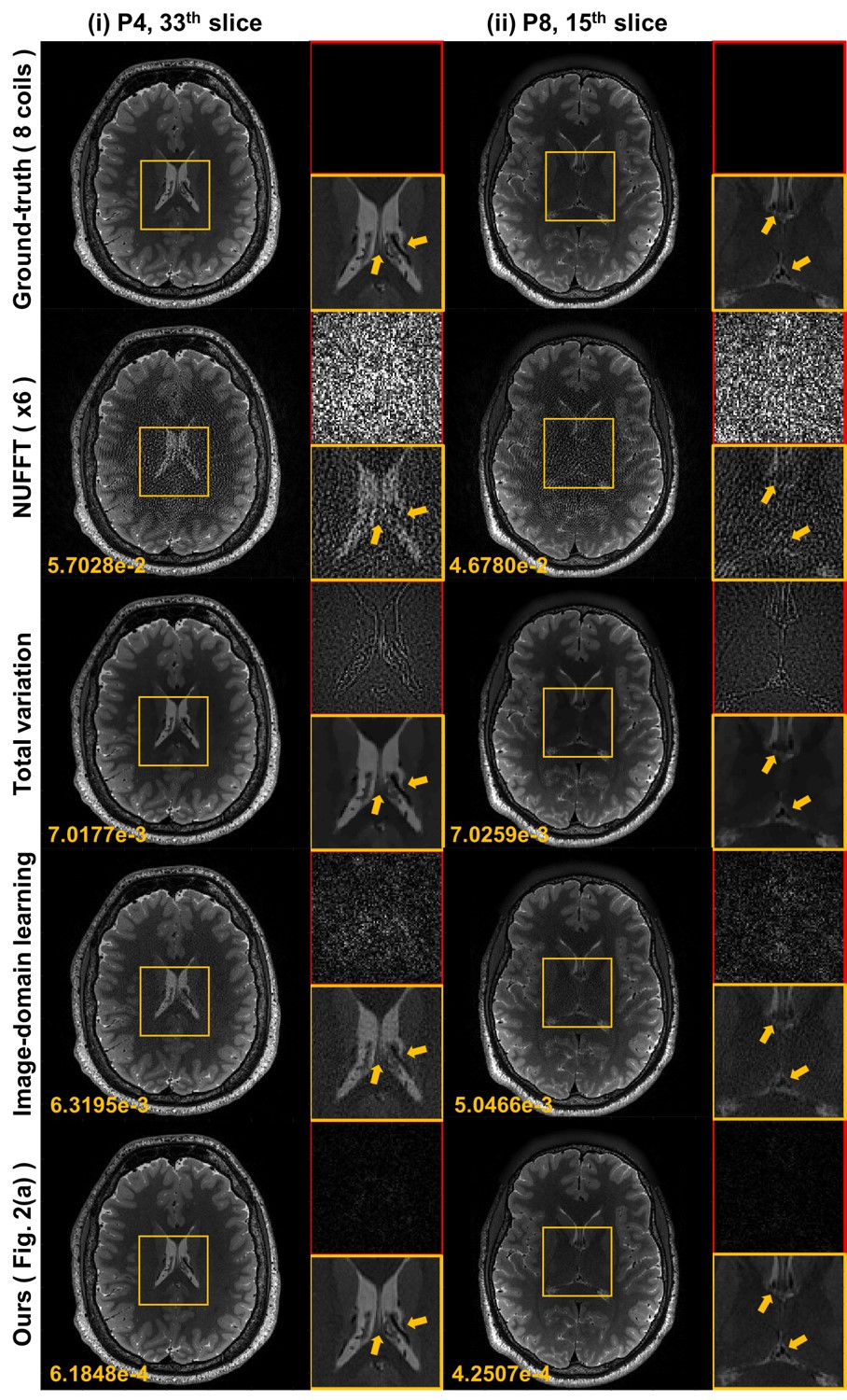}}
\caption{Reconstruction results from radial trajectory at $R=6$ in 8 coils parallel imaging. The difference images were amplified five times. Yellow and red boxes illustrate the enlarged and difference views, respectively.  The number written to the images is the NMSE value.}
\label{fig:result_radial}
\end{figure}

Fig. \ref{fig:result_radial} shows the reconstruction images from x6 accelerated radial sampling patterns using the architecture in Fig.~\ref{fig:architecture}(a) for 8 coils parallel imaging. The results for single coil are shown in Fig. S1 in Supplementary Material. 
The proposed $k$-space deep learning provided realistic image quality and preserves the detailed structures as well as the textures, but the image domain network failed to remove the noise signals and the total variation method did not preserve the realistic textures and sophisticated structures. Our method also provides much smaller NMSE values, as shown at the bottom of each Fig. \ref{fig:result_radial}  and Fig. S1 in Supplementary Material. Average PSNR, NMSE and SSIM values are shown in Table \ref{tbl:result_radial} and Table S1 in Supplementary Material for multi coils and single coil cases, respectively. 
The average values were calculated across all slices and 9 subjects. The proposed $k$-space deep learning provided the best quantitative values. 

\begin{figure}[!t] 	
\centerline{\includegraphics[width=0.9\linewidth]{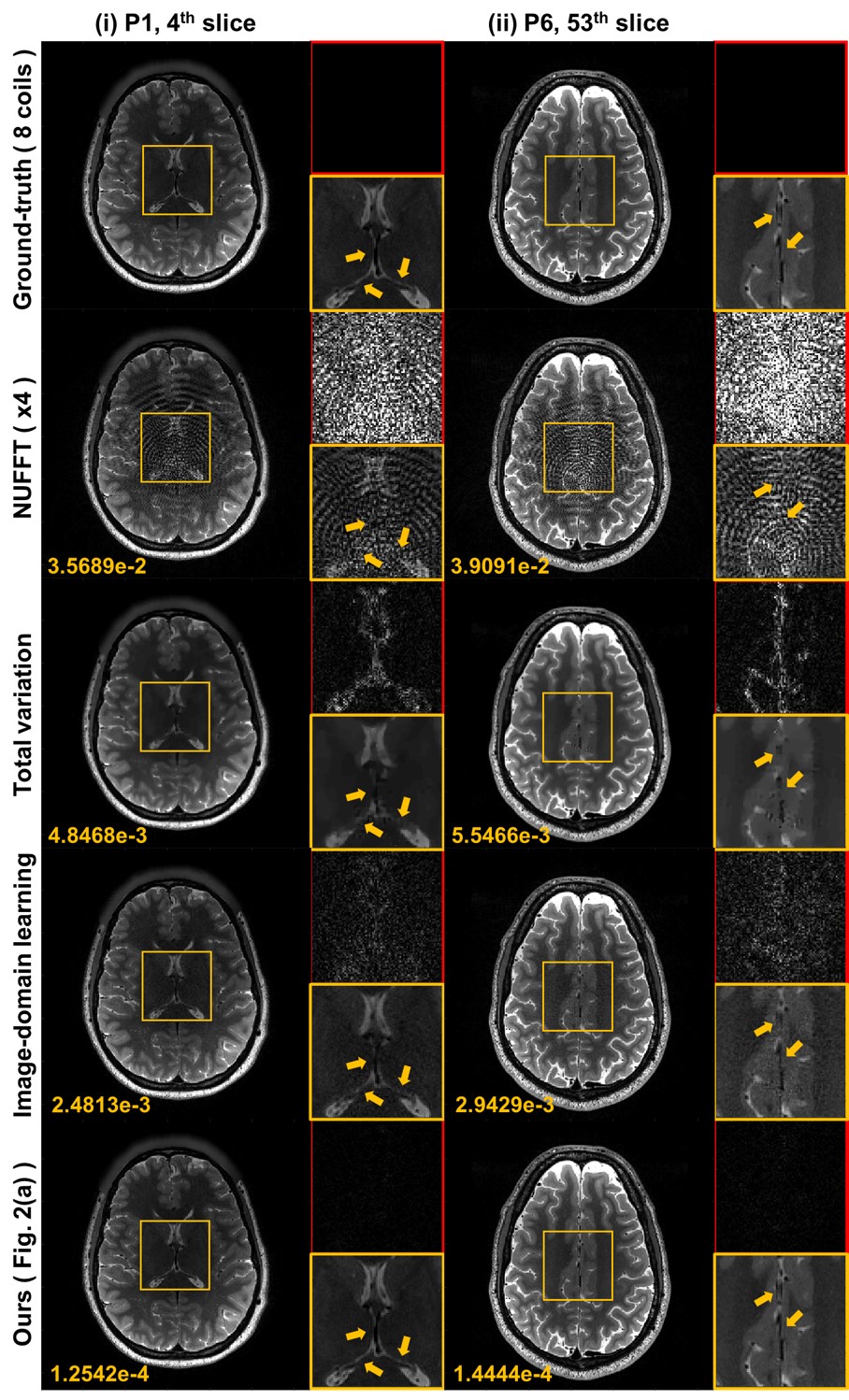}}
\caption{Reconstruction results from spiral trajectory at $R=4$ in 8 coils parallel imaging. The difference images were amplified five times. Yellow and red boxes illustrate the enlarged and difference views, respectively. The number written to the images is the NMSE value.}
\label{fig:result_spiral}
\end{figure}

\begin{table}[!t]
\centerline{
\resizebox{0.48\textwidth}{!}{
	\begin{tabular}{c|c|c|c|c}
		\hline
			\multirow{2}{*}{Metric}	& Input  	& \multirow{2}{*}{Total variation}	& Image-domain 	& Ours 	\\ 
									& ( 8 coils, $\times 6$ )	& 			& learning		& ( Fig. \ref{fig:architecture}(a) ) 	\\ \hline\hline
			PSNR [dB]				& 30.6396	& 38.2357		& 39.9140		& \textbf{50.8136}	\\
			NMSE ($\times 10^{-2}$)	& 3.6137	& 0.5998		& 0.4224		& \textbf{0.0353}	\\ 
			SSIM					& 0.5927	& 0.9528		& 0.9020		& \textbf{0.9908}	\\ \hline
	\end{tabular}
		}
}
\vspace*{0.1cm}
\caption{Quantitative comparison from radial undersampling at $R=6$ in 8 coils parallel imaging. }


\label{tbl:result_radial}
\end{table}

 Fig. \ref{fig:result_spiral} shows the reconstruction images from x4 accelerated  spiral trajectory for 8 coils parallel imaging and Fig. S2 in Supplementary Material illustrate the single coil results, respectively.
Similar to the radial sampling patterns, the proposed method provides significantly improved image reconstruction results, and the average PSNR, NMSE and SSIM values in Table \ref{tbl:result_spiral} and Table S2 in Supplementary Material also confirm that the proposed method consistently outperform other method for all patients. 

\begin{table}[!t]
\centerline{
\resizebox{0.48\textwidth}{!}{
	\begin{tabular}{c|c|c|c|c}
		\hline
			\multirow{2}{*}{Metric}	& Input  	& \multirow{2}{*}{Total variation}	& Image-domain 	& Ours 	\\ 
									& ( 8 coils, $\times 4$ )	& 			& learning		& ( Fig. \ref{fig:architecture}(a) ) 	\\ \hline\hline
			PSNR [dB]				& 30.3733	& 38.7646		& 41.0581		& \textbf{53.5643}	\\
			NMSE ($\times 10^{-2}$)	& 4.0373	& 0.6130		& 0.3404		& \textbf{0.0201}	\\ 
			SSIM					& 0.6507	& 0.9487		& 0.8881		& \textbf{0.9940}	\\ \hline
			Times (sec/slice)		& -			& 7.8818		& 0.1375		& 0.1431			\\ \hline
	\end{tabular}
		}
}
\caption{Quantitative comparison from  spiral undersampling at $R=4$ in 8 coils parallel imaging.}
\label{tbl:result_spiral}
\end{table}

To evaluate the improvements of the proposed method, a radiologist (with 9 years of experience) thoroughly reviewed the reconstructed images. For radial trajectory images, the pyramidal tract (arrows in Fig. S3(i) of the Supplementary Material), a bundle of nerve fibers conducting motor signal from the motor cortex to the brainstem or to the spinal cord, was evaluated. It was noted that high signal intensity of the pyramidal tract was exaggerated on the TV method, which could be misdiagnosed as an abnormal finding. In addition, the ability to discriminate a pair of internal cerebral veins were evaluated. While the TV and image-domain learning methods can not differentiate the two veins, the proposed method is able to show the two internal cerebral veins separately (arrows on Fig. S3(ii) in Supplementary Material). With regard to spiral trajectory images (Fig. S4 in the Supplementary Material), the TV method shows bright dot-like artifacts along several slices (Fig. S4(i)). When the small T2 hyperintensity lesions in the left frontal lobe were evaluated, the TV method fails to demonstrate the lesions. The image domain learning preserves the lesions, but the margin of the lesions is blurry in the noisy background. The small lesions are clearly depicted on the proposed method (arrows in Fig. S4(ii)). Overall, the quality of image reconstruction of the proposed method was superior to that of other methods.

%

\section{Discussion}

In order to improve the performance of ALOHA, the matrix pencil size should be significantly large, which is not possible in standard ALOHA formulation due to the large memory requirement and extensive computational burden. 
We believe that one of the important reasons that the proposed k-space deep learning provides better performance than ALOHA is that  the cascaded convolution results in much longer filter length. Nevertheless, 
with the same set of trained filters, our network can  be adapted to different input images due to the
combinatorial nature of ReLU nonlinearity. We believe that this contributes to the benefits of k-space learning over ALOHA.
Moreover, efficient signal representation in  $k$-space domain, which is the key idea of ALOHA,  can 
synergistically work with the expressivity of the neural network to enable better performance than image domain learning.

\section{Conclusion}\label{sec:conclusion}

Inspired by a link between the ALOHA and deep learning,
this paper showed that fully data-driven $k$-space interpolation is feasible by using $k$-space deep learning
 and the image domain loss function. 
The proposed $k$-space interpolation network  outperformed the
existing image domain deep learning for various sampling trajectory.
%
As the proposed $k$-space interpolation framework is quite effective and also supported by novel theory, 
so we believe that this opens a new area of researches for many Fourier imaging problems.

%
%



\pagebreak

\beginsupplement


\title{Supplementary Material for \\``$k$-Space   Deep Learning for Accelerated MRI'' }
\date{\vspace{-4ex}}

\author{Yoseob~Han$^{1}$,~
        Leonard Sunwoo$^{2}$,~
        and~Jong~Chul~Ye$^{1, *}$,~\IEEEmembership{Senior Member,~IEEE}
}	

\maketitle

%
\begin{abstract}
\setstretch{1}
In this supplement,  we provide a detailed discussion on the theoretical aspect of $k$-space deep learning.
Additional experimental results are also provided.
\end{abstract}


\setcounter{section}{0}
\section{Hankel Matrix Construction}

For simplicity, here we consider 1-D signals, but its extension to 2-D is straightforward \cite{ye2017deep}.
 In addition,  to avoid separate treatment of boundary conditions, we assume the periodic boundary condition.
Let $\fb=\begin{bmatrix} f[0] & \cdots & f[N-1]\end{bmatrix}\in \Cd^N$ be the signal vector.
Then, a single-input single-output (SISO) convolution of the input $\fb$  can be represented in a matrix form:
\begin{eqnarray}\label{eq:SISO}
\yb = \fb\circledast \overline\hb &=& \hank_d(\fb) \hb \ ,
\end{eqnarray}
where 
 $\overline \hb$ denotes the time-reversal vector of $\hb$ under periodic boundary condition, and
$\hank_d(\fb)$ is a wrap-around  Hankel matrix  defined by
 \begin{eqnarray} 
\hank_d(\fb) = 
        \begin{bmatrix}
        f[0]  &  f[1] & \cdots   &  f[d-1]    \\
        f[1]  &   f[2] & \cdots   &  f[d]   \\
         \vdots    & \vdots     &  \ddots    & \vdots    \\
        f[{N-1}]  &   f[N] & \cdots   &  f[d-2]  
        \end{bmatrix}
    \end{eqnarray}
where $d$ denotes the matrix pencil parameter.
On the other hand, multi-input multi-output (MIMO) convolution for the $P$-channel
input $\Zb=[\zb_1,\cdots,\zb_P]$ to generate $Q$-channel output
$\Yb=[\yb_1,\cdots,\yb_Q]$
can be represented  by
\begin{eqnarray}\label{eq:MIMO}
\yb_i = \sum_{j=1}^{P} \zb_j\circledast \overline\psib_i^j,\quad i=1,\cdots, Q
\end{eqnarray}
where 
$\overline\psib_i^j \in \Rd^d$ denotes the length $d$- filter that convolves the $j$-th channel input to compute its contribution to 
the
$i$-th output channel. 
By defining the MIMO filter kernel $\Psib$ as follows:
\begin{eqnarray}
\Psib = \begin{bmatrix} \Psib_1 \\ \vdots \\ \Psib_P \end{bmatrix} \, \quad \mbox{where} \quad \Psib_j =  \begin{bmatrix} \psib_1^j  & \cdots & \psib_Q^j \end{bmatrix} \in \Rd^{d\times Q}  
\end{eqnarray}
the corresponding matrix representation of the MIMO convolution is then given by 
\begin{eqnarray}
\Yb &=& \Zb \circledast \overline\Psib = \sum_{j=1}^P \hank_d(\zb_j) \Psib_j  = \hank_{d|P}\left(\Zb\right) \Psib \label{eq:multifilter}
\end{eqnarray}
where    $\overline\Psib$ is a  block structured matrix:
\begin{eqnarray}
\overline\Psib = \begin{bmatrix} \overline\Psib_1 \\ \vdots \\ \overline\Psib_P \end{bmatrix} \, \quad \mbox{where} \quad \Psib_j =  \begin{bmatrix} \overline\psib_1^j  & \cdots & \overline\psib_Q^j \end{bmatrix} \in \Rd^{d\times Q}  
\end{eqnarray}
and $\hank_{d|P}\left(\Zb\right)$ is  an {\em extended Hankel matrix}  by stacking  $P$ Hankel matrices side by side: 
\begin{eqnarray}\label{eq:ehank}
\hank_{d|P}\left(\Zb\right)  := \begin{bmatrix} \hank_d(\zb_1) & \hank_d(\zb_2) & \cdots & \hank_d(\zb_P) \end{bmatrix} \ . 
\end{eqnarray}

\section{Real and Imaginary Channel Splitting}
For a given $\fb\in \Cd^N$, 
let 
\begin{eqnarray}\label{eq:Bb}
\Bb:= \begin{bmatrix}  \re\left[\hank_d(\fb)\right]  &  \im\left[\hank_d(\fb)\right]  \\ -  \im\left[\hank_d(\fb)\right]  &  \re\left[\hank_d(\fb)\right]  \end{bmatrix}
\end{eqnarray}
and
$$\Tb= \frac{1}{\sqrt{2}}\begin{bmatrix} \Ib_N &  \Ib_N \\ \iota \Ib_N & -\iota \Ib_N \end{bmatrix} \  .$$
Then, we can easily see that $\Tb$ is an orthonormal matrix and 
\begin{eqnarray*}
\Tb^\top \Bb \Tb&=&  \begin{bmatrix} \hank_d(\fb) &  0 \\ 0 &  \hank_d^*(\fb) \end{bmatrix}  \  ,
\end{eqnarray*}
which leads to
$$\rank \begin{bmatrix}  \re\left[\hank_d(\fb)\right]  &  \im\left[\hank_d(\fb)\right]  \\ -  \im\left[\hank_d(\fb)\right]  &  \re\left[\hank_d(\fb)\right]  \end{bmatrix}$$
$$ = \rank \hank_d(\fb)+  \rank \hank_d^*(\fb)=2\rank\hank_d(\fb).$$
Therefore,
\begin{eqnarray}\label{eq:our}
\rank \begin{bmatrix}  \re\left[\hank_d(\fb)\right]  &  \im\left[\hank_d(\fb)\right]   \end{bmatrix} \leq  2\rank \hank_d(\fb),
\end{eqnarray}
One could use \eqref{eq:Bb} directly for Hankel matrix decomposition,  but this introduce additional
complication in the implementation due to the interaction between real and imaginary channels.
Thus, for simplicity, we use \eqref{eq:our}  as a surrogate for the original rank constraint.

\section{Additional Experimental Results}

\subsection{Single coil results for radial and spiral trajectories}

\begin{figure}[!hbt] 	
\centering
\centerline{\includegraphics[width=0.9\linewidth]{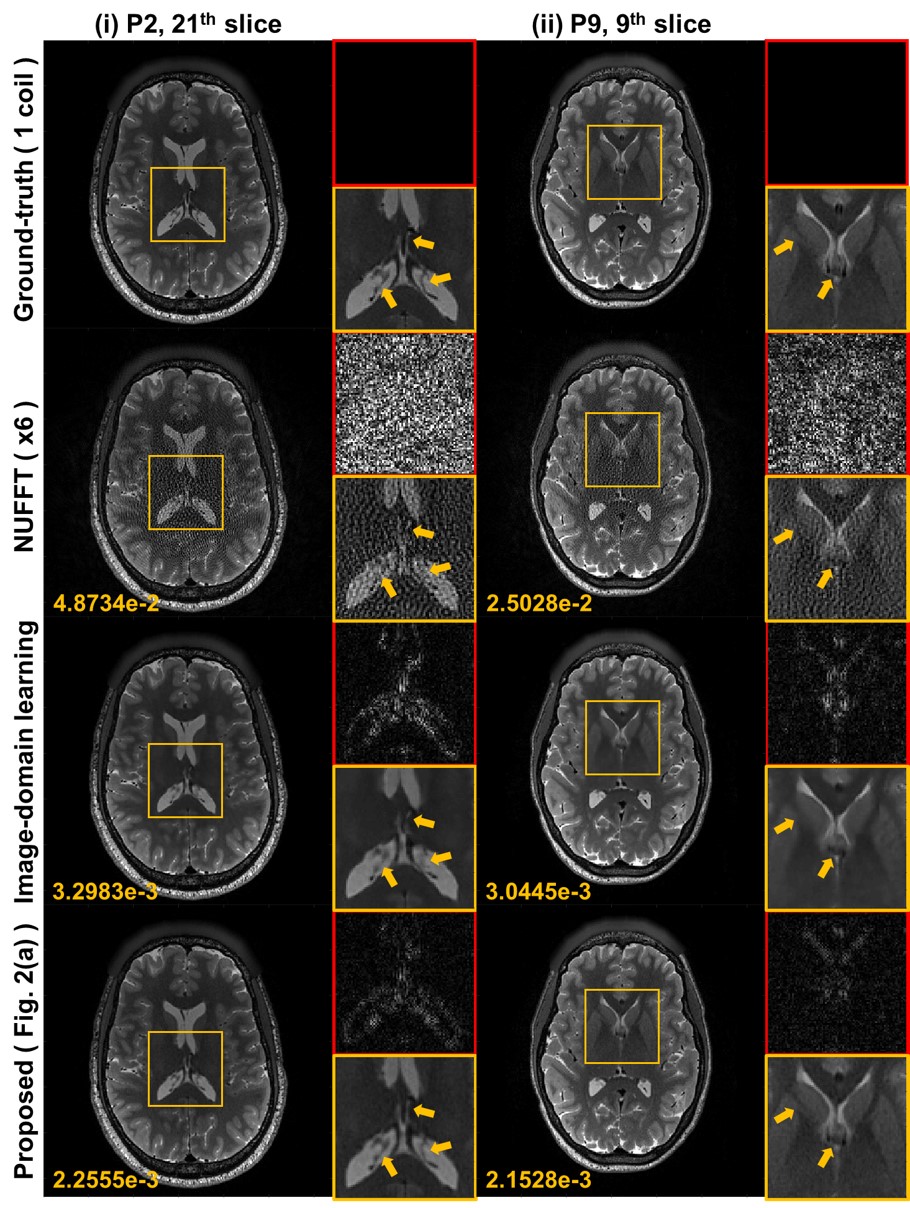}}
\caption{Reconstruction results from radial trajectory at $R=6$ in single coil imaging. The difference images were amplified five times. Yellow and red boxes illustrate the enlarged and difference views, respectively. The number written to the images is the NMSE value.}
\label{fig:result_radial_sup}
\end{figure}

The reconstruction results  from radial trajectory at $R=6$ in single coil imaging are shown in Fig.~\ref{fig:result_radial_sup}
and their quantitative comparison are given in Table~\ref{tbl:result_radial_sup}.
The proposed method outperformed the image domain approach in terms of image quality and quantitative metric except the SSIM.
In the radial sampling pattern, the SSIM value is slightly lower than the image-domain network. However,  when calculating the SSIM value within brain region, the image-domain network is 0.9741, whereas the proposed method is  ${0.9809}$, which is superior than the image-domain. 

\begin{figure}[!hbt] 	
\centerline{\includegraphics[width=0.9\linewidth]{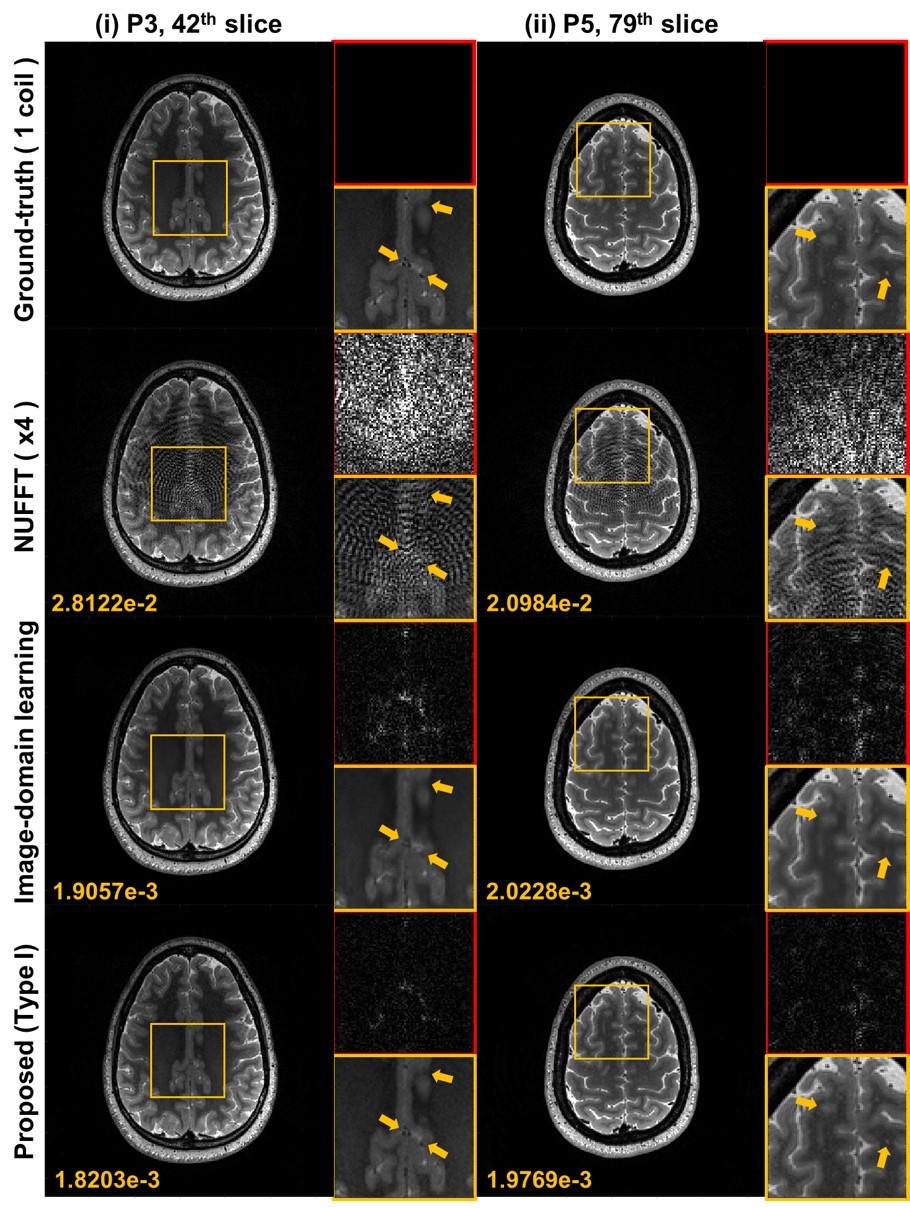}}
\caption{Reconstruction results from spiral trajectory at $R=4$ in single coil imaging. The difference images were amplified five times. Yellow and red boxes illustrate the enlarged and difference views, respectively. The number written to the images is the NMSE value.}
\label{fig:result_spiral_sup}
\end{figure}

Reconstruction results from spiral trajectory at $R=4$ in single coil imaging are shown in Fig.~\ref{fig:result_spiral_sup}
and their quantitative comparison are given in Table~\ref{tbl:result_spiral_sup}.
Similar to the radial case, the proposed method outperformed the image domain approach in terms of image quality and quantitative metric except the SSIM.
However, the SSIM value within the brain region was ${0.9872}$ by the proposed method which outperforms the image-domain SSIM value of 0.9747 in the brain region. These results suggest that the proposed method accurately recover the brain regions.

\begin{table}[!h]
\centerline{
\resizebox{0.48\textwidth}{!}{
	\begin{tabular}{c|c|c|c}
		\hline
								 \multirow{2}{*}{Metric}	& Input  	& Image-domain 			& {Ours} \\ 
															& ( 1 coil, $\times 6$ )& learning	& \multicolumn{1}{c}{( Fig. 2(a) )} \\ \hline\hline
		 						PSNR [dB]					& 30.6396	& 41.3309			& \textbf{42.8454}	\\
								NMSE ($\times 10^{-2}$)		& 3.6137	& 0.3052			& \textbf{0.2202}	\\ 
								SSIM						& 0.5927	& \textbf{0.9674}	& 0.9447	\\ \hline
								Times (sec/slice)		& -			& 0.0313			& 0.0393			\\ \hline				
	\end{tabular}
		}
}
\caption{Quantitative comparison from radial undersampling at $R=6$ in  single coil. }
\label{tbl:result_radial_sup}
\end{table}
\begin{table}[!h]
\centerline{
\resizebox{0.48\textwidth}{!}{
	\begin{tabular}{c|c|c|c}
		\hline
			\multirow{2}{*}{Metric}	& Input  	& Image-domain 			& {Ours} \\ 
									& ( 1 coil, $\times 4$ )& learning	& \multicolumn{1}{c}{( Fig. 2(a) )} \\ \hline\hline
			PSNR [dB]				& 30.3733	& 42.0571			& \textbf{45.1545}	\\
			NMSE ($\times 10^{-2}$)	& 4.0373	& 0.2727			& \textbf{0.1391}	\\ 
			SSIM					& 0.6507	& \textbf{0.9677}	& 0.9480	\\ \hline
			Times (sec/slice)		& -			& 0.0313			& 0.0393			\\ \hline
	\end{tabular}
		}
}
\caption{Quantitative comparison from  spiral undersampling at $R=4$ in  single coil.}
\label{tbl:result_spiral_sup}
\end{table}

\subsection{Radiological evaluation}

\begin{figure}[!hbt] 	
\centerline{\includegraphics[width=0.9\linewidth]{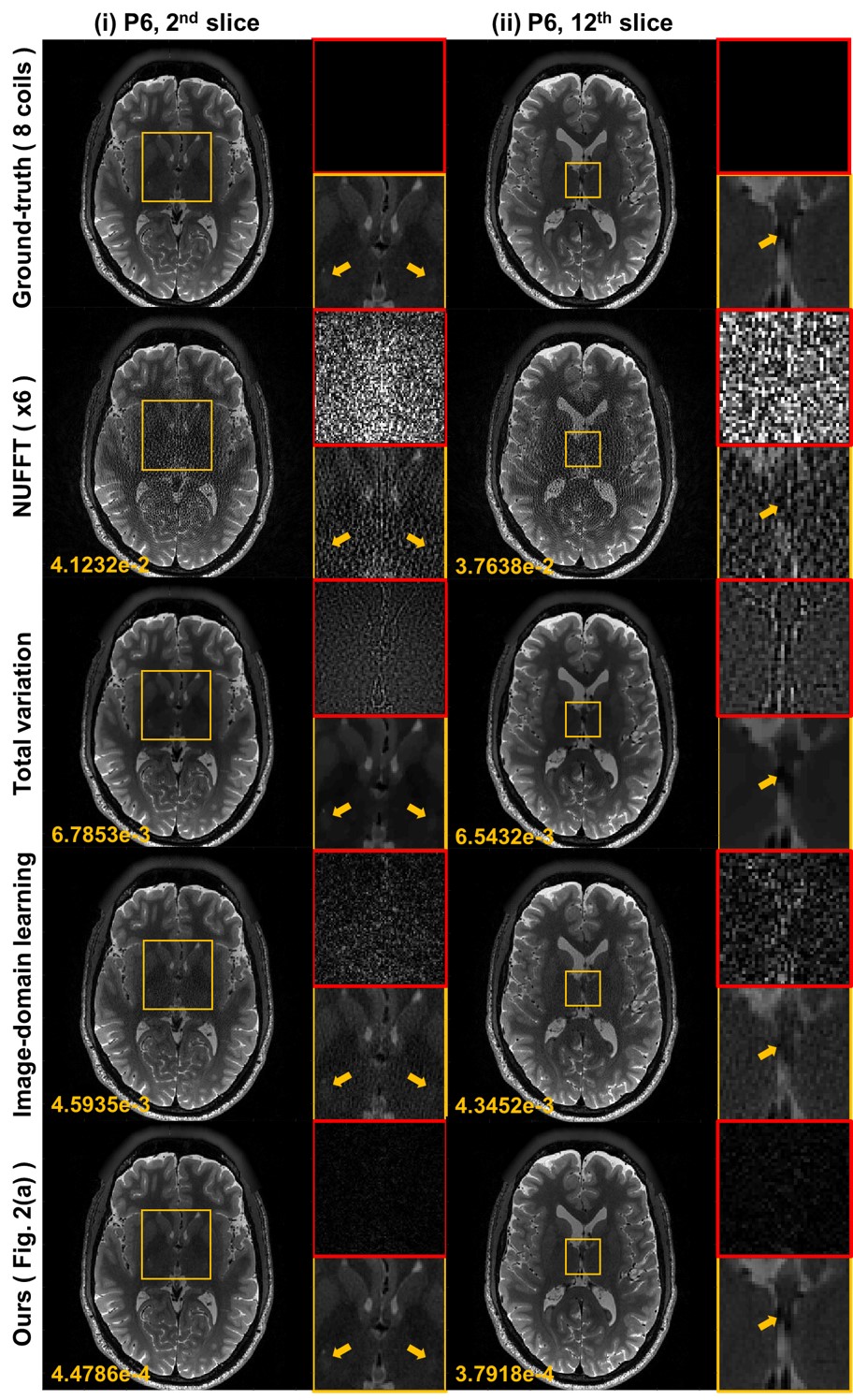}}
\caption{Reconstruction results from radial trajectory at $R=6$ in 8 coils parallel imaging. The difference images were amplified five times. Yellow and red boxes illustrate the enlarged and difference views, respectively. The number written to the images is the NMSE value.}
\label{fig:result_radial_sup_eval}
\end{figure}
\begin{figure}[!hbt] 	
\centerline{\includegraphics[width=0.9\linewidth]{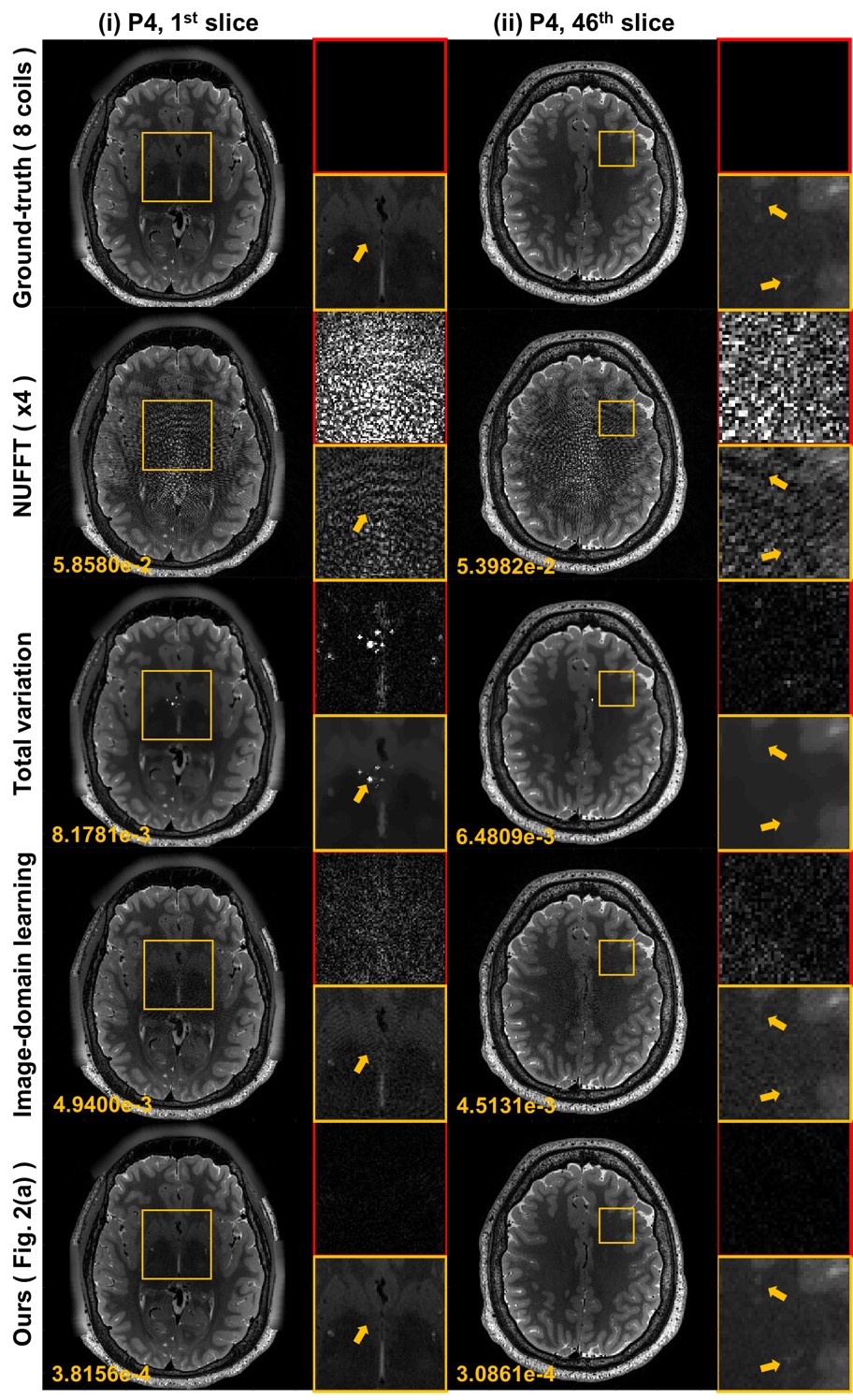}}
\caption{Reconstruction results from spiral trajectory at $R=4$ in 8 coils parallel imaging. The difference images were amplified five times. Yellow and red boxes illustrate the enlarged and difference views, respectively. The number written to the images is the NMSE value.}
\label{fig:result_spiral_sup_eval}
\end{figure}

The non-Cartesian parallel images were evaluated by an experienced radiologist, who again confirmed the superiority of 
$k$-space deep learning compared to the image domain approach.
Specifically, 
for radial trajectory results in Fig. \ref{fig:result_radial_sup_eval}, the TV method shows the watercolor-like patterns, and the image-domain learning dose not remove the streak-artifacts clearly. However, the proposed method preserved the sophisticated structure and detailed texture. In particular, yellow arrows in Fig. \ref{fig:result_radial_sup_eval}(i) indicate the pyramidal tract that is a bundle of nerve fibers to control the motor functions of the body, by conducting impulses from the motor cortex of the brain to the brainstem or spinal cord.
Although it normally shows mild hyperintensity on T2-weighted images, the brighter T2 signal intensity of the pyramidal tract on TV method compared to the ground-truth poses a risk to be mistaken for abnormal finding such as motor neuron disease, acute ischemia, or hypoglycemia.
In addition, 
yellow arrows in Fig. \ref{fig:result_radial_sup_eval}(ii) shows  internal cerebral veins.
The proposed method  clearly displays the two separate internal cerebral veins, whereas
the comparative studies such as TV and image-domain learning can not differentiate two veins.
Fig. \ref{fig:result_spiral_sup_eval} shows the reconstruction results from spiral trajectory. In the TV method in Fig. \ref{fig:result_spiral_sup_eval}(i), there are remains of the bright dot-like artifacts along several slices. Yellow arrows in Fig. \ref{fig:result_spiral_sup_eval}(ii) indicate small T2 hyperintensity lesions
 in the left frontal lobe. The proposed method clearly depicts the small sized lesions, but the TV method fails to show
 the lesions. The image-domain learning also preserves the lesions, but the margin of the lesion is blurry in the noisy background.

\subsection{Sensitivity with respect to noisy data}

\begin{figure*}[!hbt]
	\centerline{\includegraphics[width=0.8\linewidth]{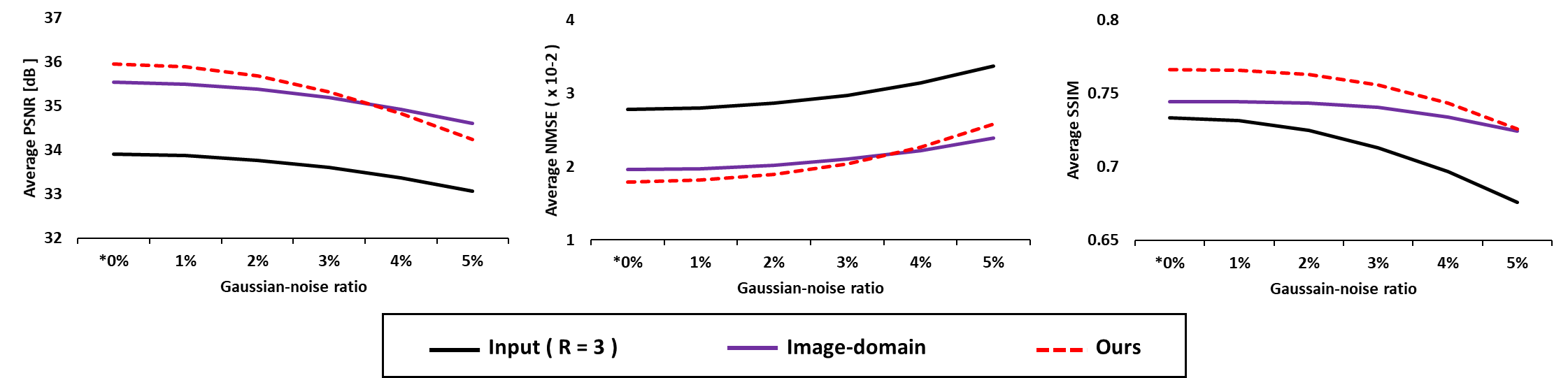}}
	\caption{Quantitative comparison with respect to noise. Gaussian noise data has been added to the input from 0\% to 5\% of  themaximum intensity of ground-truth. 0\% indicates the noise-free case.}
	\label{fig:sensitivity}
\end{figure*}

Fig.~\ref{fig:sensitivity} shows the comparison results using noisy data for the case of single coil cartesian trajectory. The noise was generated as 0\% to 5\% of the maximum intensity of ground-truth, and added to the $k$-space domain. The proposed $k$-space learning method outperforms the image-domain learning for the noise scales between 0\% and 3\%. In the noise scale beyond the 3\%, the proposed $k$-space learning method shows more errors in PSNR and NMSE than the image-domain method, but the structural similarity (SSIM) still outperforms.
However, these results were generated from the neural network trained with only noiseless data. If the noisy data are included in the training phase, the proposed method can directly learn the relationship between noisy to noiseless $k$-space mapping, which may improve the robustness.

\subsection{Effects of weighting}

\begin{figure*}[!hbt]
	\centerline{\includegraphics[width=0.6\linewidth]{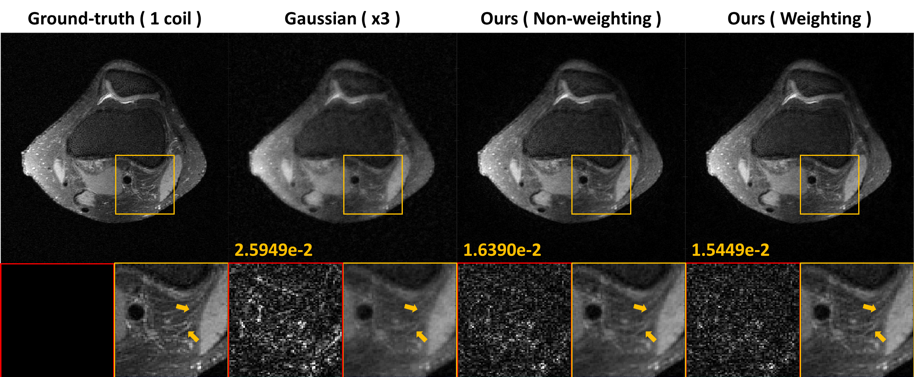}}
	\caption{Reconstruction results from Cartesian trajectory at $R = 3$ from single coil data. The difference images were amplified five times. Yellow and red boxes illustrate the enlarged and difference images, respectively. The number written in the images is the NMSE value.}
	\label{fig:weighting}
\end{figure*}

Fig.~\ref{fig:weighting} compares the reconstruction results by the non-weighted network and weighted network. The non-weighted network does not preserve the detail structures and has the blurring artifacts. However, weighted networks not only provide sharper images than weightless networks, but also provide the better quantitative performance as shown in Table~\ref{tbl:weighting}.

\begin{table}[!h]
\centerline{
\resizebox{0.48\textwidth}{!}{
	\begin{tabular}{c|c|c|c}
		\hline
									& Input  	&  Without weighting			& With weighting \\ \hline
		 						PSNR [dB]				&	33.9058	& 35.7544 & \textbf{35.9586}	\\
								NMSE ($\times 10^{-2}$)		& 2.7779	 &1.8659	& \textbf{1.7921}	\\ 
								SSIM						& 0.7338	&0.7662	& \textbf{0.7664} \\ \hline
	\end{tabular}
		}
}
\caption{Quantitative comparison from cartesian undersampling at $R=3$ in  single coil. }
\label{tbl:weighting}
\end{table}

\begin{figure*}[!hbt]
	\centerline{\includegraphics[width=0.8\linewidth]{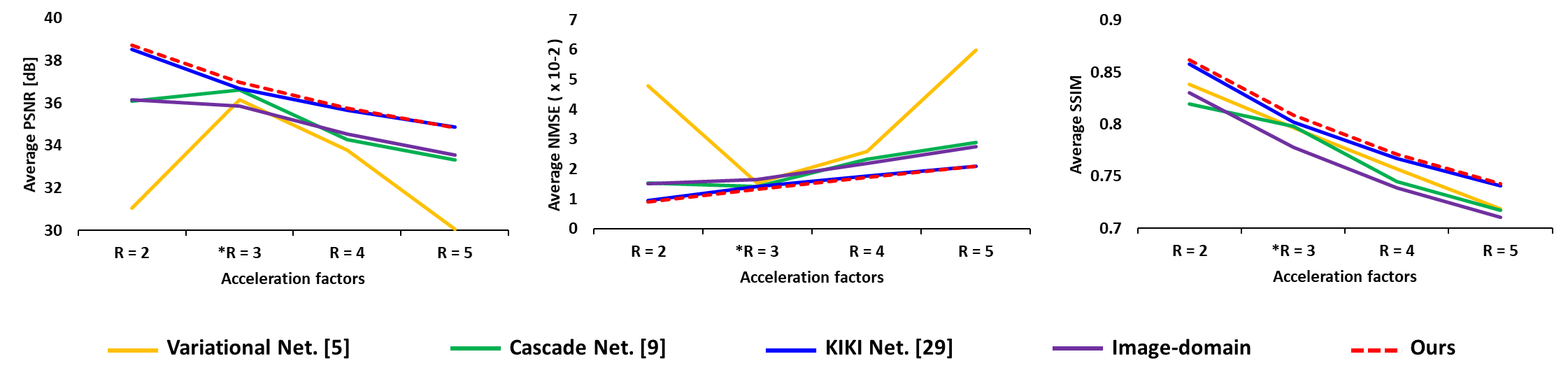}}
	\caption{Quantitative comparison with respect to various acceleration factors.}
	\label{fig:adversarial}
\end{figure*}

\subsection{Robustness to adversarial conditions}

\begin{figure}[!hbt]
	\centerline{\includegraphics[width=\linewidth]{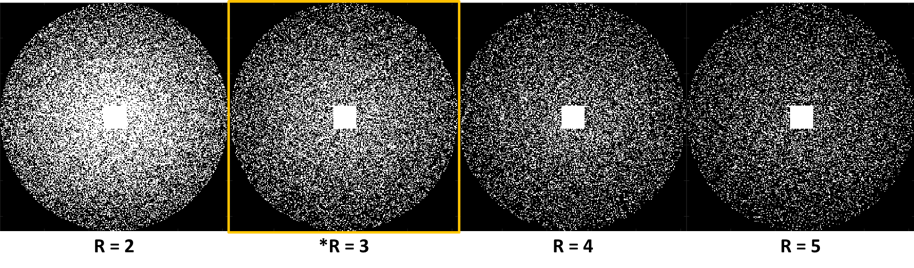}}
	\caption{Various sampling patterns. Except for a sampling pattern of $R = 3$ marked with a yellow box, no other patterns were used in the training phase.}
	\label{fig:adversarial_sampling}
\end{figure}

In Fig.~\ref{fig:adversarial}, additional experiments are performed to address the adversarial attacks or robustness of DL-based MR reconstruction. Here, the network was trained with $R=3$ acceleration factors in Figure~\ref{fig:adversarial_sampling}, which is used to reconstruction images from other acceleration factors in Figure~\ref{fig:adversarial_sampling}. The data was from cartesian trajectory with 8 coils. As shown in Figure ~\ref{fig:adversarial}, image-domain methods such as Variational Net, Cascade Net, and image-domain learning showed performance degradation in $R = 2$, but the performance was improved in Fourier-domain methods such as KIKI Net  and Ours. Overall, the Fourier-domain methods were robust than the image-domain methods with respect to various acceleration and sampling patterns that were not used during training. This again shows the advantages of the $k$-space learning.

\section{Understanding Geometry of Encoder-Decoder CNNs}

%


Although we have revealed the relations between ALOHA  and deep learning in a single image setting,
it is not clear how  these relations would translate when training is performed over multiple images.  
 Specifically, when multiple images are considered, one may suspect that
 large dimensional subspaces would be required to  approximate all of them. 
 Since the subspaces are related to sparsity of each image,  it is important to understand how would the learning be generalizable to images whose sparsity patterns may be very different. 
In addition, there are multiple skipped connection in U-Nets, which should be understood in the context of approximating all of the training data.

 \begin{figure*}[!hbt]
	\centerline{\includegraphics[width=0.9\linewidth]{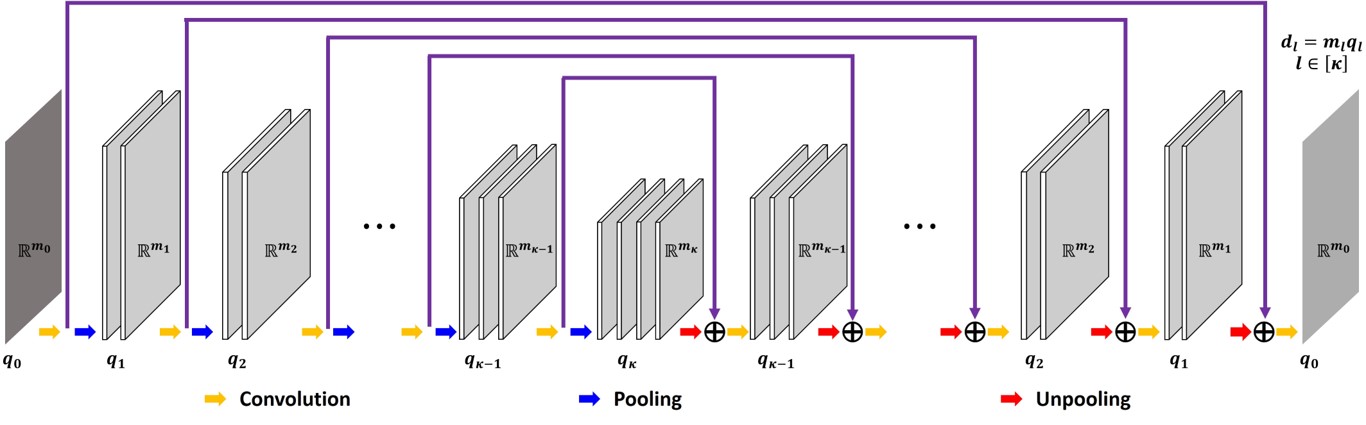}}
	\caption{An architecture of $\kappa$-layer symmetric encoder-decoder CNN with skipped connections. Here, $q_l$ denotes the number of channels at the $l$-th layer, whereas $m_l$ refers to each channel dimension, and $d_l$ represents the total
	dimension of the feature at the $l$-th layer.}
	\label{fig:network}
\end{figure*}

One may expect that the above learning from multiple images seem to have some relation to dictionary learning.
For example, 
the  sparsity prior in dictionary learning enables the selection of appropriate basis functions from the dictionary for each specific image, enabling adaptation/generalization to the specific image.


In fact, inspired by recent theoretical understanding  of neural networks, in our companion paper \cite{ye2019cnn}, we provide a unified theoretical framework  that leads to a better understanding of geometry of encoder-decoder CNNs.
Our unified mathematical framework shows that encoder-decoder CNN architecture is closely related to nonlinear basis representation
using  combinatorial convolution frames, whose expressibility increases exponentially with the network depth.
We also 
demonstrate  the importance of skipped connection 
 in terms
of expressibility,  and  optimization landscape.

While the technical details can be found in our companion paper \cite{ye2019cnn},
here we summarize our findings to make the paper self-contained.

\subsection{Expressivity}

For simplicity, consider  encoder-decoder networks in Fig.~\ref{fig:network}
which have  symmetric configuration similar to U-Net.
Specifically, the encoder network maps a given input signal $\xb\in\Xbc\subset \Rd^{d_0}$ to a 
 feature space $\zb \in \Zbc\subset \Rd^{d_\kappa}$, whereas the decoder
 takes this feature map  as an input, process it  and produce an output
$\yb \in \Ybc\subset \Rd^{d_L}$.
In this paper, symmetric configuration is considered so that
both encoder and decoder have the same number of layers, say $\kappa$;
the input and output dimensions for the encoder layer $\Ec^l$ and the decoder layer $\Dc^l$ are symmetric:
\begin{eqnarray*}
\Ec^l:\Rd^{d_{l-1}} & \mapsto& \Rd^{d_l}, \\ 
\Dc^l:\Rd^{d_{l}} &\mapsto& \Rd^{d_{l-1}}
\end{eqnarray*}
where $l\in [\kappa]$ with $[n]$ denoting the set $\{1,\cdots, n\}$;
and both input and output dimension is $d_0$.

%
%

In fact, one of the important roles of using ReLU is that it allows combinatorial basis selection such that
exponentially large number of basis expansion is feasible once the network is trained.
This is in contrast with the standard framelet basis estimation.
For example, for a given target data
$\Yb = \begin{bmatrix} \yb_{(1)} & \cdots & \yb_{(M)} \end{bmatrix}$  and the input data $\Xb= \begin{bmatrix} \xb_{(1)} & \cdots & \xb_{(M)} \end{bmatrix}$,
the estimation problem of the frame basis and its dual without
nonlinearity is optimal for the given training data,  but the network is not expressive and 
does not generalize well when the different type of input data is given.
Thus, one of the important requirements  is to allow large number of  expressions that are adaptive to the different
inputs.

Indeed,  ReLU nonlinearity serves to makes the network more expressive. For example, consider a trained two layer encoder-decoder CNN:
\begin{eqnarray}
\yb = \tilde \Bb \Lambdab(\xb)\Bb^\top\xb
\end{eqnarray}
where $\tilde \Bb\in \Rd^{d_0\times d_1}$ and $ \Bb\in \Rd^{d_0\times d_1}$ 
and $\Lambdab(\xb)$ is a diagonal matrix with 0, 1 elements that are determined by the ReLU output.
Now, the matrix can be equivalently represented by
\begin{eqnarray}\label{eq:proj}
\tilde \Bb\Lambdab(\xb)\Bb^\top = \sum_{i=1}^{d_1} \sigma_i(\xb) \tilde \bb_i \bb_i^{\top} 
\end{eqnarray}
where $\sigma_i(\xb)$ refers to the $(i,i)$-th diagonal element of $\Lambdab(\xb)$.
Therefore, depending on the input data $\xb\in \Rd^{d_0}$,  $\sigma_i(\xb)$ is either 0 or 1 so that a maximum  $2^{d_1}$ distinct
configurations of the matrix can be represented using \eqref{eq:proj}, which 
is significantly more expressive than using  the single representation with the frame and its dual.
This  observation can be generalized as shown in the following theorem which can be found in \cite{ye2019cnn}.

\begin{theorem}\label{thm:decexp}\cite{ye2019cnn}
Let
\begin{eqnarray}
 \tilde\Upsilonb^l= \tilde\Upsilonb^l(\xb) :=  \tilde\Upsilonb^{l-1} \tilde\Lambdab^{l}(\xb) \Db^{l} ,~ \label{eq:UD0} \\
   \Upsilonb^{l}=   \Upsilonb^{l}(\xb) := \Upsilonb^{l-1} \Eb^{l} \Lambdab^{l}(\xb)  ,~ \label{eq:UE0}
\end{eqnarray}
with $ \tilde\Upsilonb^0(\xb) =\Ib_{d_0}$ and $ \Upsilon^{0}(\xb) =\Ib_{d_0}$, and
\begin{eqnarray}
 \Mb^l=\Mb^l(\xb) :=\Sb^{l}\Lambdab_S^{l}(\xb)\label{eq:M0} \\
\tilde \Mb^l=\tilde \Mb^l(\xb) := \tilde \Lambdab^{l}(\xb)\tilde \Sb^{l} \label{eq:tM0} 
\end{eqnarray}
where $\Eb^l$ and $\Db^l$ denote the matrix representation of  the $l$-th layer  convolution layer
for encoder and decoder, respectively;
  $\Lambdab^l(\xb)$ and  $\tilde\Lambdab^l(\xb)$ refer to the  diagonal matrices from ReLU at the $l$-th layer encoder and decoder, respectively,
which have 1 or 0 values;  $\Lambdab_S^l(\xb)$  refers to a similarly defined diagonal matrices from ReLU at the  $l$-th skipped branch
of encoder. 
Then, the following statements are true.

1) Under ReLUs, an encoder-decoder CNN  without skipped connection can be represented by
\begin{eqnarray}\label{eq:feed}
\yb =  \tilde\Bb(\xb)\Bb^{\top}(\xb)\xb  = \sum_i \langle \xb, \bb_i(\xb) \rangle \tilde \bb_i(\xb) 
\end{eqnarray}
where
\begin{eqnarray}\label{eq:Bcx}
 \Bb(\xb) = \Upsilonb^\kappa(\xb)&,& \tilde \Bb(\xb) = \tilde\Upsilonb^\kappa(\xb)
\end{eqnarray}
Furthermore,  the maximum number of available linear representation is given by
\begin{eqnarray}\label{eq:nproj}
 N_{rep} = 2^{\sum_{i=1}^{\kappa}d_i-d_\kappa},\quad
 \end{eqnarray}
 
 2) An encoder-decoder CNN  with skipped connection under ReLUs is given by
 \begin{eqnarray}\label{eq:skipnet}
 \yb = \tilde\Bb^{skp}(\xb)  \Bc^{skp \top}(\xb)\xb  = \sum_i  \langle \xb, \bb_i^{skp}(\xb) \rangle \tilde \bb_i^{skp}(\xb) 
 \end{eqnarray}
 where
 $$ \Bb^{skp}(\xb) := $$
 \begin{eqnarray}\label{eq:Bcxskip}
 \begin{bmatrix} \Upsilonb^\kappa & \Upsilonb^{\kappa-1}\Mb^\kappa &  \Upsilonb^{\kappa-2}\Mb^{\kappa-1} & \cdots & \Mb^1\end{bmatrix}
\end{eqnarray}
 $$\tilde \Bb^{skp}(\xb) := $$
 \begin{eqnarray}\label{eq:tBcxskip}
 \begin{bmatrix}\tilde \Upsilonb^\kappa & \tilde\Upsilonb^{\kappa-1}\tilde \Mb^\kappa & \tilde \Upsilonb^{\kappa-2}\tilde \Mb^{\kappa-1} & \cdots & \tilde \Mb^1\end{bmatrix}
\end{eqnarray}
Furthermore,  the maximum number of available linear representation is given by
\begin{eqnarray}\label{eq:nproj2}
 N_{rep} = 2^{\sum_{i=1}^{\kappa}d_i-d_\kappa}\times 2^{\sum_{i=1}^\kappa s_k}
 \end{eqnarray}
\end{theorem}

This implies that  the number of representation  increase exponentially with the network
depth, which again confirm the expressive power of the neural network.
Moreover, the skipped connection also significantly increases the expressive power of the encoder-decoder CNN.
Another important consequence of Theorem~\ref{thm:decexp} is that
 the input space $\Xbc$ is partitioned into the maximum $N_{rep}$ non-overlapping
regions so that inputs for each region shares the same linear representation.

Due to the ReLU, one may wonder whether the cascaded convolutional interpretation of the convolutional
framelet still holds.
A close look of the proof  in \cite{ye2019cnn}  reveals that  this is still the case.
Specifically, ReLUs  provides spatially varying mask to the convolution filter so that the net effect
 is a convolution with the  the spatially varying filters originated from masked version of convolution filters \cite{ye2019cnn}.
This results in  a spatially variant cascaded convolution, and only change in the interpretation
  is that the basis and its dual are composed of  {\em spatial variant} cascaded convolution filters.
 Furthermore, the ReLU works to diversify the convolution filters by masking out the various filter coefficients. It is believed that this is another source of expressiveness from the same set of convolutional filters. 



\end{document}